\documentclass[10pt,journal,compsoc]{IEEEtran}
\usepackage{xspace}
\usepackage{url}
\newcommand*{\eg}{\emph{e.g.}\@\xspace}
\newcommand*{\etc}{\emph{etc}\@\xspace}
\newcommand*{\ie}{\emph{i.e.}\@\xspace}

\newcommand*{\vs}{\emph{v.s.}\@\xspace}

\newcommand{\history}{Open-VCLIP\xspace}
\newcommand{\system}{Open-VCLIP++\xspace}

\usepackage{microtype}
\usepackage{graphicx}
\usepackage{subfigure}
\usepackage{booktabs} 

\usepackage{amsmath}
\usepackage{makecell}
\usepackage{multirow}
\usepackage{enumitem}
\usepackage{bbding}
\usepackage{amsmath}
\usepackage{MnSymbol}
\usepackage{wasysym}

\usepackage{amsthm}
\usepackage{colortbl}
\usepackage[dvipsnames]{xcolor}
\definecolor{green_im}{rgb}{0.1, 0.55, 0.3}
\newcommand{\drop}[1]{\textcolor{red}{\scriptsize{$\downarrow$#1}}}
\newcommand{\rise}[1]{\textcolor{green_im}{\scriptsize{$\uparrow$#1}}}

\DeclareMathOperator*{\argmin}{arg\,min}

\usepackage{subfigure}
\usepackage{parskip}

\usepackage[ruled,boxed,commentsnumbered]{algorithm2e}
\usepackage{algorithmic}

\newtheorem{theorem}{Theorem}[section]

\newtheorem{lemma}[theorem]{Lemma}

\usepackage[numbers,sort&compress,comma]{natbib}

\usepackage[pagebackref=true,breaklinks=true,letterpaper=true,colorlinks,citecolor=ForestGreen, bookmarks=false]{hyperref}

\usepackage[capitalize,noabbrev]{cleveref}

\begin{document}
\title{Building an Open-Vocabulary Video CLIP Model with Better Architectures, Optimization and Data}

\author{Zuxuan Wu$^*$,
        Zejia Weng$^*$,
        Wujian Peng,
        Xitong Yang,  
        Ang Li, \\
        Larry S. Davis,~\IEEEmembership{Fellow,~IEEE}
        and Yu-Gang Jiang
\IEEEcompsocitemizethanks{\IEEEcompsocthanksitem Z. Wu, Z. Weng, W. Peng, Y-G. Jiang are with School of Computer Science, Fudan University. \protect  E-mail: \{zxwu, zjweng20, wjpeng22, ygj\}@fudan.edu.cn 
\IEEEcompsocthanksitem X. Yang is with Meta AI. \protect Email: xyang35@meta.com
\IEEEcompsocthanksitem A. Li is with Simular AI.  \protect Email: me@angli.ai
\IEEEcompsocthanksitem  LS Davis is with the Department
of Computer Science, University of Maryland, College Park.  \protect E-mail: lsd@umiacs.umd.edu
\IEEEcompsocthanksitem $^{*}$ Equal contributions.
}}

\markboth{Journal of \LaTeX\ Class Files,~Vol.~14, No.~8, August~2015}%
{Shell \MakeLowercase{\textit{et al.}}: Bare Demo of IEEEtran.cls for Computer Society Journals}

\IEEEtitleabstractindextext{%
\begin{abstract}
Despite significant results achieved by Contrastive Language-Image Pretraining (CLIP) in zero-shot image recognition, limited effort has been made exploring its potential for zero-shot video recognition. This paper presents \system, a simple yet effective framework that adapts CLIP to a strong zero-shot video classifier, capable of identifying novel actions and events during testing. \system minimally modifies CLIP to capture spatial-temporal relationships in videos, thereby creating a specialized video classifier while striving for generalization. We formally demonstrate that training \system is tantamount to continual learning with zero historical data. To address this problem, we introduce Interpolated Weight Optimization, a technique that leverages the advantages of weight interpolation during both training and testing. Furthermore, we build upon large language models to produce fine-grained video descriptions. These detailed descriptions are further aligned with video features, facilitating a better transfer of CLIP to the video domain.
Our approach is evaluated on three widely used action recognition datasets, following a variety of zero-shot evaluation protocols. The results demonstrate that our method surpasses existing state-of-the-art techniques by significant margins. Specifically, we achieve zero-shot accuracy scores of 88.1\%, 58.7\%, and 81.2\% on UCF, HMDB, and Kinetics-600 datasets respectively, outpacing the best-performing alternative methods by 8.5\%, 8.2\%, and 12.3\%. We also evaluate our approach on the MSR-VTT video-text retrieval dataset, where it delivers competitive video-to-text and text-to-video retrieval performance, while utilizing substantially less fine-tuning data compared to other methods. Code is released at {\small\url{https://github.com/wengzejia1/Open-VCLIP}}.
\end{abstract}

\begin{IEEEkeywords}
Zero-Shot Recognition, Video Recognition, CLIP, Language Models
\end{IEEEkeywords}}

\maketitle
 
\IEEEdisplaynontitleabstractindextext

\IEEEpeerreviewmaketitle

\IEEEraisesectionheading{\section{Introduction}\label{sec:introduction}}

\IEEEPARstart{Z}{ero-shot} learning aims to recognize novel unseen categories during inference without having seen them during training. It is a non-trivial problem yet particularly useful in real-world scenarios where manual labels are extremely difficult and expensive to collect.
Among a plethora of work studying zero-shot learning~\cite{zellers2017zero,brattoli2020rethinking,xu2017transductive}, CLIP~\cite{radford2021learning} recently emerges as a strong zero-shot learner. Leveraging web-scale image and text pairs in a contrastive manner, CLIP delivers remarkable zero-shot image recognition
results across a broad spectrum of tasks, ranging from image segmentation~\cite{wang2022cris,ghiasi2022scaling}, image editing~\cite{zheng2022bridging,crowson2022vqgan}, vision-to-text retrieval~\cite{luo2022clip4clip,portillo2021straightforward,wang2022internvideo}, out-of-distribution detection~\cite{esmaeilpour2022zero,shu2023CLIPood,ming2022delving}, \etc.

While significant improvements have been made in zero-shot image recognition, exploring CLIP for zero-shot video action recognition remains under-explored. Adapting CLIP, initially designed for image tasks, to the video domain is extremely challenging, particularly in the zero-shot scenarios. Video analysis requires capturing temporal dynamics——although one could combine frame-based predictions with temporal pooling~\cite{wang2021actionclip}, it has been found that enhancing CLIP with dedicated temporal modeling components on top of off-the-shelf image models produces better results~\cite{ni2022expanding}. However, this improvement comes at the cost of reduced generalization, as it requires fine-tuning CLIP and may lead to overfitting due to the smaller scale of video datasets compared to the original CLIP training data. As a result, the zero-shot ability of CLIP diminishes over the course of fine-tuning. 

\begin{figure}[t]
    \centering
    \includegraphics[width=0.5\textwidth]{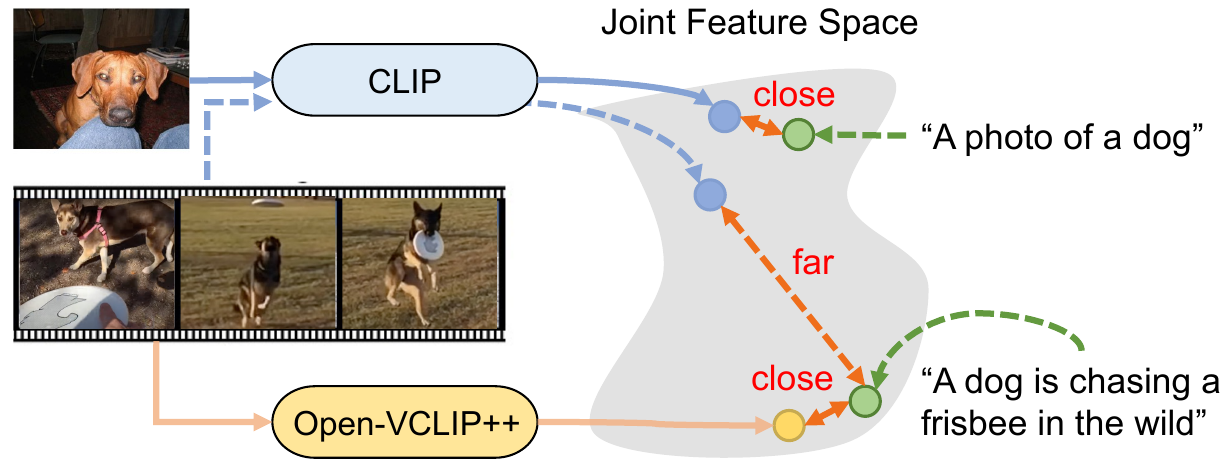}
    \caption{{\color{black} While CLIP has shown impressive results for zero-shot image recognition, it cannot effectively recognize novel actions in videos. This paper aims to transform CLIP to a strong zero-shot video classifier with minimal modifications.} }
    \label{fig:pipeline}
\end{figure}

It is worth noting that the fine-tuning process transforms CLIP, an image classifier renowned for its robust zero-shot capabilities, to a specialized video understanding model. This inspires us to explore a balanced approach, striking a middle ground between generality and specialization. Our goal is to adapt a pretrained CLIP model to the video domain such that the derived model can predict seen actions and events effectively through temporal modeling and recognize novel categories in a zero-shot manner like CLIP as well. Notably, we discover from a theoretical perspective that this problem is essentially a continual learning problem, which aims to adapt a pretrained model to new tasks with less forgetting on previous knowledge. Traditional continual learning focuses on achieving satisfactory performance across all previously encountered tasks with access to historical data~\cite{hu2021opin,balaji2020taskonomycl,shin2017continual}. However, our situation presents unique challenges due to the unavailability of raw training data for CLIP. Moreover, our objective extends beyond typical continual learning: we hope to transform CLIP into an effective zero-shot video learner, one that exhibits robust generalization to unknown video actions and events, rather than exactly preserving its knowledge for image tasks, which again is challenging without access to the training data of CLIP.

With this in mind, we study how to construct an open-vocabulary video CLIP by leveraging the pretrained weights that are publicly available. We build upon CLIP with minimal modifications so as to derive a video action recognition model that not only captures temporal information among frames but also generalizes well to unseen actions and events during testing. To enhance the continual learning-based training process, we propose a novel training method called \textit{Interpolated Weight Optimization}, which regularizes the fine-tuning process of the CLIP model by establishing a connection between the original CLIP model and the current model. This prevents the derived model from drifting away from CLIP, which we argue is beneficial for generalization. Moreover, beyond training, we also establish connections between derived optima along the optimization trajectory during testing to further enhance generalization. Furthermore, we propose to improve the fine-tuning process by generating pseudo captions with language models.
These derived captions are more descriptive and can enrich the textual diversity in the video training data whose labels usually exist in the forms of words or simple sentences describing the category of interest. As a result, \system is able to achieve a better transition of CLIP from the image domain into the video domain.

We conduct extensive experiments to evaluate the performance of \system. In particular, we extend CLIP to the video domain by leveraging Kinetics-400 as a proxy dataset and then evaluate zero-shot recognition results on UCF-101, HMDB-51 and Kinetics-600 under different evaluation settings. \system demonstrates state-of-the-art zero-shot video action recognition performance, offering 88.1\%, 58.7\%, 81.2\% zero-shot accuracy on UCF, HMDB and Kinetics-600 respectively, surpassing alternative methods by  8.5\%, 8.2\% and 12.3\%. Furthermore, \system also achieves the best trade-off between close-set and zero-shot performance across all the benchmarks. We also evaluate the zero-shot video-text retrieval performance on MSR-VTT dataset, and results suggest that \system is an efficient approach that can substantially improve the video-text retrieval capability of CLIP model, achieving 39.0\% text-to-video recall and 37.8\% video-to-text recall, without any additional manually labeled video text pairs.

A preliminary version of this paper appeared in ~\cite{weng2023transforming}. 
The present paper includes a complete literature review on open-vocabulary video models; an updated solution that regularizes the learning of video-text alignment with the help of pseudo captions to prevent knowledge forgetting; more thorough experiments and discussion to demonstrate the compatibility of \system; additional experiments on image classification tasks to demonstrate our method is able to maintain the original knowledge of CLIP for image tasks as well.

\section{Related Work}
\textbf{Zero-shot Video Action Recognition.} Zero-shot video action recognition requires deep neural networks to classify novel categories that are not seen during training. This capability proves valuable in real-world scenarios where acquiring data and their corresponding labels is challenging. 
Early research primarily focuses on how to represent actions properly. Various approaches are introduced, including utilizing manually-crafted attributes \cite{liu2011recognizing,zellers2017zero}, object features \cite{jain2015objects2action,gao2019know} to represent actions. Additionally, researchers explore the use of word embedding to represent actions as textual semantic representations \cite{brattoli2020rethinking,xu2017transductive}. Recent interest has shifted towards pretraining models on large-scale vision-text datasets, demonstrating remarkable success in zero-shot image classification \cite{radford2021learning,jia2021scaling}. There is also a plethora of work leveraging knowledge acquired from large-scale pretrained vision-language models for downstream tasks in a zero-shot manner \cite{wang2022cris,ghiasi2021open,ghiasi2022scaling}. Despite the wealth of research in zero-shot image understanding, the domain of zero-shot video classification remains less explored. While ActionCLIP \cite{wang2021actionclip} and XCLIP \cite{ni2022expanding} experiment with zero-shot settings for action recognition, they largely adopt image-based strategies for zero-shot learning and overlook the challenge of model forgetting during CLIP adaptation. In contrast, our goal is to explicitly construct a robust open-vocabulary zero-shot video classifier by regularizing the fine-tuning process.

\textbf{Continual Learning.}
Continual learning, a field dedicated to training models on a sequence of tasks while avoiding the catastrophically forgetting of previously acquired knowledge, encompasses  strategies that can be split into three main categories: memory-based, expansion-based, and regularization-based methods. Memory-based methods typically employ a replay buffer to store past examples or relevant information like gradients, as demonstrated in previous successful continual learning work \cite{farajtabar2019orthogonal,hu2021opin,balaji2020taskonomycl}. However, in our specific context, direct application of these approaches is not feasible due to the absence of historical data. Expansion-based methods, exemplified by Progressive Neural Networks (PNN) \cite{pnn}, progressively expand the network architecture over time to maintain performance on earlier tasks. Closest to our approach, perhaps are regularization-based methods \cite{yin2020sola} such as EWC \cite{ewc}, which add a regularization term to the optimization that enforces new model weights to remain close to their original values. Nonetheless, it is essential to note that methods like EWC still require historical data for calculating the Fisher information matrix. Unlike traditional continual learning, our approach aims to transfer knowledge, specifically the ability to perform zero-shot learning, from a previous image task to video tasks, without access to any historical data at all.

\textbf{Captioning with Large Language Models.} 
Visual captioning serves as a pivotal link connecting the fields of computer vision and natural language processing. Recent advancements, driven by multimodal pretraining and the integration of vision-language understanding and generation in approaches like BLIP~\cite{li2022blip,li2023blip2}, have yielded remarkable performance gains in zero-shot image captioning.  With the advent of diverse large language models (LLMs)~\cite{GPT4,touvron2023llama}, researchers have extended these models to help extract more nuanced descriptive information. One prevalent approach involves treating LLMs as agents responsible for aggregating various types of information describing images and videos. Prior work has effectively utilized LLMs to enhance vision descriptions by aggregating fine-grained details, including objects, attributes, and frame-level captions~\cite{rotstein2023fusecap,wang2022language,ghosh2023captext}.
Another type of LLM-based captioning treats LLMs as agents providing feedback. Notably, ChatCaptioner~\cite{zhu2023chatgpt} employs ChatGPT to continuously ask contextually relevant questions about images to BLIP-2, resulting in enriched image descriptions. These methods demonstrate the potential of LLM-equipped captioning models to provide more comprehensive descriptions for both images and videos. In contrast to this line of research, we not only use LLM-equipped captioning models to generate detailed video captions, but also utilize the captions to further enhance the alignment capabilities of the multimodal models.

\section{Preliminary: Video Action Recognition Using CLIP}
\label{sec:pre}

Recognizing actions in videos is a fundamental yet challenging task, typically demanding intensive model training on large-scale datasets. Inspired by the success of contrastive language-image pretraining (CLIP)~\cite{radford2021learning}, recent research has proposed to fine-tune the  well-trained CLIP model on the specific video dataset of interest and has achieved state-of-the-art results~\cite{xu2021videoclip,wu2022transferring}.

In the pursuit of adapting CLIP for video action recognition (VCLIP), a prevalent approach~\cite{arnab2021vivit,bulat2021space,bertasius2021space,xing2022svformer} is to extend the image encoder to encompass temporal dynamics within videos and, in the meanwhile, align the video representation with the corresponding text representations of its corresponding labels like ``chasing frisbee''. Specifically, let $V\in \mathcal{V}_B$ and $T\in \mathcal{T}_B$ denote a video clip and its corresponding action label described in textual prompts, and the primary objective during fine-tuning is to maximize the similarity:
\begin{align}
    \text{sim}(v, t) = \frac{\langle v, t\rangle}{\|v\| \|t\|}, \;\;\; v=f_{\theta_B^V}(V), \; t=f_{\theta_B^T}(T),
\end{align}
if $V$ and $T$ represent the same video. Here, $\mathcal{V}_B$ denotes the video dataset for task $B$, $\mathcal{T}_B$ denotes the corresponding label set, $f_{\theta_B^V}$ denotes the visual encoder, and $f_{\theta_B^T}$ denotes the text encoder. It is worth noting that 
the text encoder is usually frozen during training~\cite{ilharco2022patching,thengane2022clip},
thus the fine-tuning stage primarily concentrates on the optimization of the visual encoder for adaptation to the video domain. For the sake of brevity, the superscript will be omitted in the subsequent paragraphs.

\section{Our Approach}

This section presents our approach, which comprises three key components: (1) transforming an image-based CLIP model into a video CLIP model VCLIP to capture spatial-temporal relationships within videos, (2) applying regularization techniques during fine-tuning such that the resulting model is capable of recognizing novel actions and events, and (3) generating detailed video captions to improve alignment between the visual and textual modalities.

\subsection{Constructing VCLIP for Video Understanding}\label{al:temporalattend}
The core of action recognition is to model temporal relationships among video frames. However, the image-based CLIP model lacks the capability to aggregate temporal features, and thus is unsuitable for video tasks. Consequently, we devote to enhancing the original CLIP model by incorporating temporal modeling abilities to facilitate a more effective transition from image-based to video-based tasks.

A straightforward way for temporal modeling is to add additional temporal networks, which however results in an increase in parameters, leading to higher computational costs and making it harder for weight interpolation that significantly enhances zero-shot capabilities, as will be described later.
Fortunately, we observe that the self-attention layer in vision transformers is quite scalable,  operating on image frame patches as:
\begin{align}
    y_{s,t} = \text{Softmax}(\frac{q_{s,t} K_t^\text{T}}{\sqrt{d}}) V_t,
\end{align}
where $d$ is the dimension of vectors, $q_{s,t}$ denotes the query vector associated with the $s$-th token in the $t$-th frame, $K_t^\text{T}$ denotes the transpose of the matrix composed of key vectors in the $t$-th frame, and $V_{t}$ represents the matrix consisting of value vectors in the $t$-th frame. It is obvious that each token exclusively acquires information from its corresponding frame. To address this limitation, we expand the temporal attention view for every self-attention layer  to facilitate the aggregation of global temporal information, achieved through the stacking of self-attention layers. The revised self-attention layer operates as follows:
\begin{align}
    y_{s,t} = \text{Softmax}\left(\frac{q_{s,t} [K_{(t-1)\sim(t+1)}]^\text{T}}{\sqrt{d}}\right) [V_{(t-1)\sim(t+1)}].
\end{align}
At this time, each image patch captures information not only from its own frame but also from neighboring frames. The special modification lies in $[K_{(t-1)\sim(t+1)}]$ and $[V_{(t-1)\sim(t+1)}]$ which refers to the Key/Value matrix be composed with key/value vectors belonging to not only the $t$-th frame, but also the adjacent frames. This subtle adjustment helps the model to gain better ability for temporal information aggregation, while fitting our following algorithm since no extra parameters are added.

\subsection{Training Open-VCLIP} \label{trainopenvclip}
 We now present the training method with a carefully designed optimization strategy for improved zero-shot video recognition. We start by formulating the problem from its original purpose and then derive its approximation which leads to a challenging continual learning problem with zero historical data. To address this problem, we introduce a novel regularization-based technique for optimizing VCLIP, which we term as \textit{Interpolated Weight Regularization} (IWR). Additionally, we integrate the Stochastic Weight Averaging (SWA) model-averaging approach to further enhance the generalizability of our model.

\subsubsection{Problem Definition}
While fine-tuning the CLIP model produces decent results for close-set video classification~\cite{wang2021actionclip,ni2022expanding}, its performance on unseen categories is limited---might be worse than the original CLIP model as shown in \cref{table:baseline-compare}. Below, we will delve into the process of creating a robust \textit{open-vocabulary} video model, building upon a pretrained CLIP model.

Formally, our goal is to obtain the optimal vision encoder $f_{\theta_U}$ that satisfies:
\begin{align}
\theta_U = \arg\min_{\theta} L(\theta;D_U).
\label{eq:universal}
\end{align}
Here, $D_U = \left\{\mathcal{V}_U, \mathcal{T}_U\right\}$ is a \emph{hypothetical} universal dataset that contains all possible videos and their corresponding text descriptions (\ie, action labels). $L$ denotes the loss function defined on the video-text pairs.
Optimizing such an objective directly is infeasible, but an alternative exists: we can approximate it by training on a comprehensive dataset that comprises a sufficiently wide range of video and text data.

\subsubsection{An Approximation Equivalent to Continual Learning} \label{problem_approximate}
We first consider a video action recognition dataset, denoted as $D_B = \left\{\mathcal{V}_B, \mathcal{T}_B \right\}$, used for fine-tuning. Unfortunately, despite the abundant video data  
in $\mathcal{V}_B$ providing a good approximation of of $\mathcal{V}_U$, the associated text space is considerably restricted, confined by the number of annotated categories (\eg, $|\mathcal{T}_B|=400$ for Kinetics-400~\cite{kay2017kinetics}). Consequently, the video representation is prone to overfitting to the highly skewed text space, and the zero-shot capability of the CLIP model deteriorates throughout the fine-tuning process. On the other hand, the size of the image training dataset for CLIP is sufficiently large to approximate $D_U$, yet a domain gap exists between the image and video spaces.
Even worse, this dataset is a private dataset and only the CLIP model weights $\theta_A$ are accessible for fine-tuning.

Considering these advantages and limitations, our aim is to utilize \textit{both} $\theta_A$ and $D_B$ to construct a strong open-vocabulary model. We postulate that $\theta_A$ should  encompass valuable information from the large-scale image dataset on which CLIP was initially trained. Following this intuition, we notice that the initial VCLIP model (without any fine-tuning) $\theta_A$ is indeed an optimal solution to a large-scale video dataset $D_{A}$ with a diverse text space $\mathcal{T}_{A}$ (\cref{lemma1}).

\begin{lemma} Suppose the image CLIP model was trained on an image-text dataset $D_{\bar{A}} =\{\mathcal I_{\bar{A}},\mathcal T_{\bar{A}}\}$ with $N$ examples. Then, there exists a diverse video-text dataset $D_A=\{\mathcal V_A, \mathcal T_A\}$ containing $N$ examples where the video CLIP model with original CLIP parameters $\theta_A$ is optimal. 
\begin{proof} We denote the large-scale visual-language pretraining dataset used to train CLIP as $D_{\bar{A}}=\{\mathcal I_{\bar{A}},\mathcal T_{\bar{A}}\}$. To bridge the gap between image and video domains, we extend images in $D_{\bar{A}}$ by repeating frames to create a set of static videos $\mathcal V_{A}=\{\mathcal V_i=[\mathcal I_{\bar{A}i}, \mathcal I_{\bar{A}i}, ..., \mathcal I_{\bar{A}i}]\}^N$ and let $\mathcal T_{A}=\mathcal T_{\bar{A}}$. Videos in $D_A=\{\mathcal V_A, \mathcal T_A\}$ are static  and do not provide any additional information for prediction. As a result, $\theta_A$ which offers the best results for $D_{\bar{A}}$ is also optimal on $D_A$.
\end{proof}\label{lemma1}
\end{lemma}

Although $D_A$ is unknown, it is helpful to understand our problem. Given the fact that $\theta_A$ is an optimal solution in a large scale dataset $D_A$, a natural idea for approximating the universal objective is to combine both datasets in the derivation, although $D_A$ is unknown in reality.

Following this idea, \cref{eq:universal} is transformed into:
\begin{align}
\arg\min_{\theta} L(\theta;D_A) + L(\theta;D_B),
\label{eq:estimation}
\end{align}
where $D_A$ is completely unknown and only $D_B$ is present. However, we have the optimal solution $\theta_A$ on $D_A$. In that case, the formulation becomes equivalent to \emph{continual learning}, \ie, continually training the model on a new dataset while preserving its performance on historical data so as to achieve sufficient generalizability.

\begin{figure*}[t]
    \centering
    \includegraphics[width=0.95\textwidth]{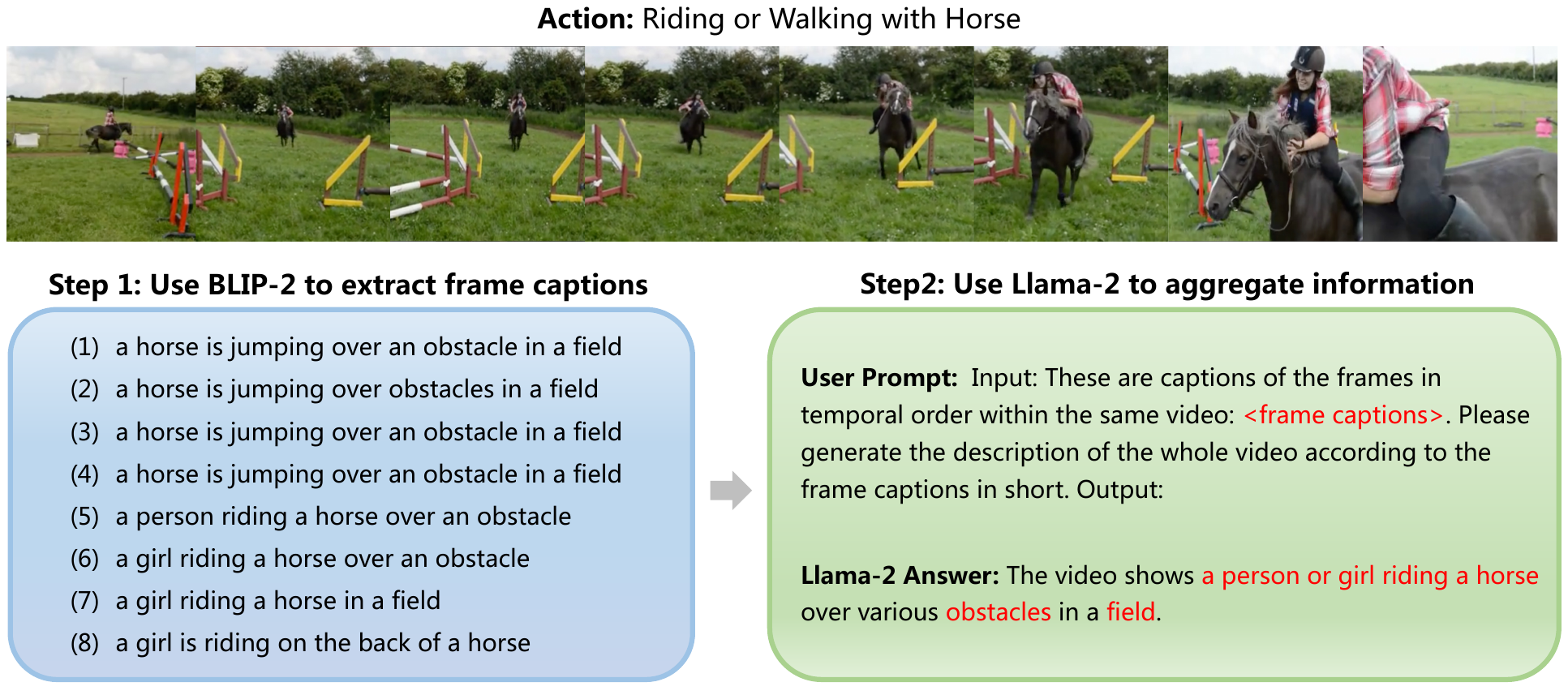}
    \caption{ We show the pipeline of generating video-level captions. We use BLIP-2 to obtain frame-level descriptions, which are further aggregated into video-level captions with Llama-2. As shown in the figure,  the video contains a single action label ``riding or walking with horse''. Instead, the caption generated by our pipeline not only includes a more accurate action description ``riding a horse'', but also encompasses additional information about the actor and the scene, such as ``a person or a girl'', ``over various obstacles'', ``in a field''.}
    \label{fig:caption_pipeline}
\end{figure*}

\subsubsection{Interpolated Weight Regularization}
Although numerous methods have been developed to investigate continual learning, the majority of them rely on the preservation of historical data or information, a strategy that is inapplicable in our context. This is primarily because we lack access to $D_A$ at all.
Motivated by empirical results in~\cite{ilharco2022patching,wortsman2022robust}, we propose to lowering the first loss term in \cref{eq:estimation} by introducing an \textit{optimization-free} weight interpolation strategy:
\begin{equation}
\begin{aligned}
&\theta=\lambda \theta_A + (1-\lambda)\theta_B,
\label{eq:patch}
\end{aligned}
\end{equation}
where $\lambda$ is a trade-off hyperparameter. \cref{eq:patch} is referred to as model patching~\cite{ilharco2022patching}, which is commonly used between two converged model weights, \ie, $\theta_B$ is trained separately on $D_B$ only. A potential risk of this method is that we have no explicit optimization on the curve fitting performance of the final patched model, \ie, the patch may be severely underfitting to $D_B$ or sensitive to the trade-off parameter $\lambda$.

To address this issue, we propose Interpolated Weight Regularization as part of the training procedure, which regularizes the loss of the interpolated weights on $D_B$. Given that the $\lambda$ is a hyperparameter that may vary, instead of optimizing a single point, we look for a solution such that the patched model's performance \emph{w.r.t.} a range of interpolation coefficients are optimized. This is achieved by sampling a balancing coefficient during training. The final optimization objective becomes:
\begin{equation}
    \argmin_{\theta_B} \ \mathcal L = L(\theta_B; D_B) + \beta {L(\alpha \theta_A + (1-\alpha) \theta_B; D_B)},
    \label{eq:final_linear_connect}
\end{equation}
where $\alpha \sim \texttt{U}(0, \lambda)$. The $(0,\lambda)$ interval corresponds to the low-level region between the interpolated and end weights. $\beta$ is the trade-off coefficient for regularization and it is set as $\beta=C\frac{1}{1-\alpha}$ in practice where $C$ is a constant value. The loss can be optimized by calculating its derivative as follows:
\begin{align}
    \frac{d\mathcal L}{d\theta} &= \left.\frac{dL}{d\theta}\right|_{\theta=\theta_B} + \beta(1-\alpha)\left.\frac{dL}{d\theta}\right|_{\theta=\alpha\theta_A+(1-\alpha)\theta_B}\\
    &= \left.\frac{dL}{d\theta}\right|_{\theta=\theta_B} + C\left.\frac{dL}{d\theta}\right|_{\theta=\alpha\theta_A+(1-\alpha)\theta_B}~.
    \label{eq:final}
\end{align}
After we obtain the optimal $\theta_B$, \cref{eq:patch} is applied to achieve the final VCLIP model weights.

\subsubsection{Stochastic Weight Averaging}

Improved zero-shot ability indicates good generalization. We further introduce an upgrade to the above method by applying Stochastic Weight Average (SWA) on the interpolated weights along the training process to find the ``flat minimum'', which refers to a region in the weight space where the loss function is relatively flat and the test error is relatively low according to \cite{izmailov2018averaging}. \cref{eq:swa} showcases our solution:
\begin{align}
    \sum_i^N \frac{{\lambda\theta_A+(1-\lambda)\theta_i}}{N} = \lambda \theta_A + (1-\lambda)\, \boxed{\frac{1}{N} \sum_i^N \theta_i} ,
    \label{eq:swa}
\end{align}
where $\theta_i$ refers to the $i$-th set of parameters we select during the training process. As \cref{eq:swa} shows, the moving average of the interpolated weights equals to interpolating the moving average of the updated weights, showing the order of the SWA and the weight interpolation is interchangable. So, in practice, we maintain the moving average of the model's weights during training and do weight interpolation in the end. By averaging model weights from the optimization trajectories, we will get a robust model with better generalization to unseen classes.

\subsection{Expanding Text Space for Better Alignment.} 

Up till now, we  have introduced how to transfer CLIP to the video domain via fine-tuning on the video action recognition dataset, denoted as $D_B = \left\{\mathcal{V}_B, \mathcal{T}_B \right\}$. However, the extremely limited number of annotated action categories (\eg, $|\mathcal{T}_B|=400$ for Kinetics-400~\cite{kay2017kinetics}) results in overfitting and catastrophic forgetting. \cref{trainopenvclip} proposes to address the forgetting problem from the aspect of continual learning (\ie, a better optimization strategy during training), and here, we further explore how can we expand the text space for action recognition dataset for improved results.  We propose to use an image captioning model combined with a large language model to obtain fine-grained video descriptions. Based on this, we align the video features with more detailed captions to combat overfitting, and thus ensure a better transfer to the video domain.

In particular, we introduce a video captioning method that automatically creates detailed descriptions for videos. 
Instead of using off-the-shelf video captioning models, which 
rely heavily on down-stream fine-tuning and lack the ability to generate text for unseen scenarios~\cite{wang2022language}, we build upon BLIP-2~\cite{li2023blip2} that excels in zero-shot visual captioning, an ideal candidate for our task. Specifically, we sample 16 frames for each video in a sparse manner, and then, each frame will be fed into BLIP-2~\cite{li2023blip2} for caption generation.

While BLIP-2 produces decent captions, the descriptions are frame-level sentences without temporal coherence. To improve video-text alignment for a better transfer of CLIP, we merge frame-level captions into video-level texts, which entails extracting key information among different frames, removing redundant information, and summarizing in a concise manner. To this end, we build upon a large language model, Llama-2~\cite{touvron2023llama}, serving as an aggregator of frame-level descriptions so as to generate cohesive and complete textual representations. This is achieved by designing appropriate prompts as inputs to Llama-2. In our implementation, we set up the prompt for the system role in Llama-2 as follows:
\begin{center} 
\emph{``Always answer in one sentence.''}
\end{center}
to constrain the answer of the model to be concise. Then we set up the prompt for the user role in Llama-2 as follows: 

\begin{center} 
\emph{``Input: These are captions of the frames in a sequential order within the same video: \textbf{ \textless ~ frame captions \textgreater}. Please summarize the whole video according to the frame captions in short. Output: The video shows''},
\end{center}

where the ``\emph{{ \textless frame captions \textgreater}}'' corresponds to frame descriptions extracted by BLIP-2. The integration of BLIP-2 and Llama-2 harnesses the strengths of both models, empowering the framework to generate detailed video descriptions. We denote the newly generated video text information with $\hat{\mathcal{T}}_B$:
\begin{align}
\hat{\mathcal{T}}_B = \text{Llama-2}(\text{BLIP-2}({\mathcal{V}_B})).
\label{eq:newcap}
\end{align}

These diverse textual descriptions can be further used for better alignment. Specifcically, we use the contrastive loss, akin to CLIP pretraining, to further align video features with the caption features. {\color{black} For example, given a video depicting ``riding or walking with horse'', the original loss aligns the video features with ``a video of [riding or walking with horse]''. In contrast, the new loss additionally aligns video features with ``The video shows a person or girl riding a horse over various obstacles in a field''.}
Now deriving the optimal $\theta^{*}$ on dataset $D_B$ follows:
\begin{align}
L(\theta^{*};D_B) = L(\theta^{*}; {\mathcal{V}_B, \mathcal{T}_B}) + \gamma \cdot L(\theta^{*}; {\mathcal{V}_B, \hat{\mathcal{T}}_B}),
\label{eq:finaltotalloss}
\end{align}
where $\gamma$ is a trade-off coefficient. 
Such a process promotes the generalization capabilities of the CLIP model while concurrently reducing overfitting.

The training and inference algorithm for \system is summarized in \cref{alg:example_training} and \cref{alg:example_testing}.

\begin{algorithm}[h]\small
   \caption{Training}
   \label{alg:example_training}
\begin{algorithmic}
   \STATE {\bfseries Input:} Model $f_{\theta}$, Step = 0, Data $D=\{V_i, y_i, \hat{y}_i\}^{N}$ in which $y_i$ is video action, and $\hat{y}_i$ is video caption. 
   \STATE {\bfseries Require:} SWA begins at $T$ step with a cycle length $c$. Param of SWA model ${\theta_{\texttt{SWA}}}$. Counting flag $l=0$. Model Param $\theta$ is initialized by the CLIP Param $\theta_\texttt{CLIP}$. Hyper-param $\gamma$, R, $\beta$. Learning rate $\delta$.\\
   \REPEAT
   \STATE Step $\leftarrow$ Step + 1 \\
   \STATE Sample  $\{V_i, y_i, \hat{y}_i\}^M \subseteq D$,
   \STATE Let: $\widetilde{V} \leftarrow \{V_i \}^M$, $\widetilde{y} \leftarrow \{y_i \}^M$, $\widetilde{\hat{y}} \leftarrow \{\hat{y}_i \}^M$  \\
   \quad \\
   \tcp{normal supervision loss.}
   \STATE $L(f_{\theta}) \leftarrow  \frac{1}{M} \sum_{i=1}^M [L(\widetilde{V}_i,\widetilde{y}_i, \theta) + \gamma \cdot L(\widetilde{V}_i,\widetilde{\hat{y}}_i, \theta)]$\\ 
   \quad \\
   \tcp{interpolation regularization.}
   \STATE Sample $\alpha \sim \texttt{Uniform(0, R)}$
   \STATE Initialize $\widetilde{\theta} \leftarrow  \alpha \cdot \theta_\texttt{CLIP} + (1-\alpha) \cdot \theta$ 
   \STATE $L(f_{\widetilde{\theta}}) \leftarrow  \frac{1}{M} \sum_{i=1}^{M} [L(\widetilde{V}_i,\widetilde{y}_i, \widetilde{\theta}) + \gamma \cdot L(\widetilde{V}_i,\widetilde{\hat{y}}_i, \theta)]$ \\
   \tcp{update model with combined loss.}
   $\theta \leftarrow  \theta - \delta \nabla_{\theta} (L(f_{\theta}) + \beta \cdot L(f_{\widetilde{\theta}})) $ \\
   \quad \\
   \tcp{stochastic weight average.}
   \IF{$\text{Step} > T$ and $\text{mod}(\text{Step} - T, c) == 0$}
   \STATE ${\theta_{\texttt{SWA}}} \leftarrow ({\theta_{\texttt{SWA}}}*l+\theta) / ({l+1})$
   \STATE $l \leftarrow l + 1$
   \ENDIF
   \UNTIL{converge}
\end{algorithmic}
\end{algorithm}

\begin{algorithm}[h]\small
   \caption{Inference}
   \label{alg:example_testing}
\begin{algorithmic}
   \STATE {\bfseries Input:} Testing Set $D=\{V_i\}^{N}$, Model Param ${\theta_{\texttt{SWA}}}$, Original CLIP Param ${\theta_{\texttt{CLIP}}}$.
   \STATE {\bfseries Require:} Set Interpolation Ratio as $\lambda$\\
   \STATE {\bfseries Initialize:} $\theta_{\texttt{FINAL}} = \lambda * \theta_\texttt{CLIP} + (1-\lambda) * {\theta_{\texttt{SWA}}}$ \\
   \STATE $\text{Predict} = f(D; \theta_{\texttt{FINAL}})$
\end{algorithmic}
\end{algorithm}

\begin{table*}[t!]
\caption{Zero-shot classification performance for various algorithms on UCF and HMDB with different protocols (see \cref{evaluation_protocol} ). }
    \vspace{-0.1in}
\label{table:sota-compare}
\begin{center}
\resizebox{.65\textwidth}{!}{
\begin{sc}
\begin{tabular}{lccccc}
\toprule
\multirow{2}{*}{Method}  & \multirow{2}{*}{encoder} & \multicolumn{2}{c}{UCF} & \multicolumn{2}{c}{HMDB} \\
\cmidrule(lr){3-4}\cmidrule(lr){5-6}
& & EP1 & EP2 & EP1 & EP2 \\ 
\midrule
GA \cite{mishra2018generative} & C3D & 17.3$\pm$1.1 & - & 19.3$\pm$2.1 & -\\
TARN \cite{bishay2019tarn} & C3D & 19.0$\pm$2.3 & - & 19.5$\pm$4.2 & - \\
CWEGAN \cite{mandal2019out} & I3D & 26.9$\pm$2.8 & - & 30.2$\pm$2.7 & -  \\
TS-GCN \cite{gao2019know} & GLNet  & 34.2$\pm$3.1 & - & 23.2$\pm$3.0 & -  \\
PS-GNN \cite{gao2020learning} & GLNet & 36.1$\pm$4.8 & - & 25.9$\pm$4.1 & - \\
E2E \cite{brattoli2020rethinking}  & R(2+1)D  & 48.0 & 37.6  & 32.7 & 26.9 \\
DASZL \cite{kim2021daszl} & TSM & 48.9$\pm$5.8 & - &  - & - \\
ER \cite{chen2021elaborative} & TSM & 51.8$\pm$2.9 & - & 35.3$\pm$4.6 & - \\
ResT \cite{lin2022cross} &ResNet101 & 58.7$\pm$3.3 & 40.6 & 41.1$\pm$3.7 & 34.4 \\
\midrule 
ActionCLIP \cite{wang2021actionclip}  & ViT-B/16&  - & 69.5 &  - & 50.5\\
Text4Vis \cite{wu2022transferring} & ViT-L/14 & 85.8$\pm$3.3 & 79.6 &  58.1$\pm$5.7 & 49.8 \\
\midrule
 \makecell{\multirow{3}{*}{\system}} & ViT-B/32 & 87.4$\pm$2.4 & 79.9 & 62.8$\pm$4.1 & 51.1  \\
 & ViT-B/16 & 90.0$\pm$1.7 & 84.1 & 65.8$\pm$4.3& 54.7  \\
 & ViT-L/14 &93.2$\pm$2.0 & 88.1 & 69.7$\pm$4.0 & 58.7 \\
\bottomrule
\end{tabular}
\end{sc}
}
\end{center}
\end{table*}

\begin{table}[h!]
\caption{We compare the zero-shot classification performance with X-CLIP on UCF and HMDB dataset with Evaluation Protocol 3.}
    \vspace{-0.1in}
\label{table:protocol3}
\begin{center}
\begin{sc}
\begin{tabular}{lcccr}
\toprule
Method  & encoder & UCF & HMDB \\
\midrule
\makecell{X-CLIP~\cite{ni2022expanding}}  & ViT-B/16& 72.0$\pm$2.3 & 44.6$\pm$5.2\\
\midrule
\makecell{\multirow{3}{*}{\system}} & ViT-B/32 & 79.7$\pm$1.2 & 51.4$\pm$0.3\\
& ViT-B/16 & 83.9$\pm$0.6 & 55.6$\pm$1.4 \\
& ViT-L/14 & 87.8$\pm$0.9 & 59.6$\pm$0.6  \\
\bottomrule
\end{tabular}
\end{sc}
\end{center}
\end{table}

\begin{table}[ht!]
\caption{Comparisons of zero-shot video action recognition performance of different algorithms on Kinetics-600.}
    \vspace{-0.1in}
\label{table:zsl-kinetics-600}
\begin{center}
\begin{sc}
\begin{tabular}{lcccc}
\toprule
Method  & encoder & Top-1 Acc & Top-5 Acc \\
\midrule
\makecell{ER~\cite{chen2021elaborative}} & - & 42.1$\pm$1.4 & 73.1$\pm$0.3 \\
\makecell{X-CLIP~\cite{ni2022expanding}} & ViT-B/16 & 65.2$\pm$0.4 & 86.1$\pm$0.8\\
\makecell{Text4Vis~\cite{wu2022transferring}}& ViT-L/14 & 68.9$\pm$1.0 & - \\
\midrule
 \makecell{\multirow{3}{*}{\system}} &ViT-B/32 & 69.5$\pm$0.7 & 91.7$\pm$0.0    \\
 & ViT-B/16 & 73.4$\pm$0.8 & 93.4$\pm$0.1  \\
 & ViT-L/14 & 81.2$\pm$0.7 & 96.4$\pm$0.2 \\
\bottomrule
\end{tabular}
\end{sc}
\end{center}
\end{table}

\section{Experiments}
We present in this section our experimental evaluation. We compare our \system method with state-of-the-art video models and perform ablation studies to reveal the characteristics of our method.

\subsection{Experimental Setup}
\subsubsection{Datasets}

In our experiments, we use the following four datasets:

\textbf{Kinetics-400\&600:} The Kinetics-400 \cite{kay2017kinetics} and Kinetics-600 \cite{carreira2018short} are large-scale datasets curated for human action recognition, comprising 400 and 600 action classes, respectively. Each dataset consists of video clips extracted from YouTube videos with an average duration of 10 seconds. The Kinetics-600 dataset augments the Kinetics-400 dataset by incorporating an additional 220 categories. These new categories offer a valuable resource of evaluating the zero-shot performance of models that have been solely trained on the Kinetics-400 dataset.

\textbf{UCF-101:} The UCF-101 dataset \cite{soomro2012ucf101} comprises 13,320 videos from 101 action categories and is widely utilized for human action recognition. Each video is a short clip captured from real-world scenarios, depicting a specific human action, with an average duration of 7.21 seconds. Officially, the dataset provides three distinct training and testing split files.

\textbf{HMDB-51:} The HMDB-51 dataset \cite{kuehne2011hmdb} contains 6,849 clips distributed across 51 action categories, each containing a minimum of 101 clips. Three official training/testing splits are offered for model evaluation. To maintain a balance of samples across categories, 1,746 videos are left ``unused'' in each spilt. This ensures that each category contains 70 samples for training and 30 for testing. This consistent number of testing samples enables a fair evaluation.

We utilize the Kinetics-400 dataset for training and conduct testing on UCF-101, HMDB-51, along with a subset of the Kinetics-600 dataset. Testing on datasets with little to no overlap with the training data can reflect the model's realistic and comprehensive zero-shot capabilities. More details regarding the evaluation protocols are described below.

\subsubsection{Evaluation protocols} \label{evaluation_protocol}
\textbf{UCF-101\&HMDB-51:} Following \cite{brattoli2020rethinking} and \cite{ni2022expanding}, we employ the following three protocols to evaluate the zero-shot capabilities on these two datasets.
\begin{itemize}[itemsep=0pt,topsep=0pt]
    \item Evaluation Protocol 1 (\textit{EP1}): We randomly select half of the classes from the test dataset, \ie, 50 from UCF and 25 from HMDB. This process is repeated ten times, and we report the average results for each dataset \cite{brattoli2020rethinking}.
    \item Evaluation Protocol 2 (\textit{EP2}): We evaluate the model on the complete datasets, \ie, assessing performance across all 101 UCF classes and all 51 HMDB classes \cite{brattoli2020rethinking}. 
    \item Evaluation Protocol 3 (\textit{EP3}): We conduct testing on the three official splits and report the average top-1 accuracy together with the standard deviation \cite{ni2022expanding}.
\end{itemize}

\textbf{Kinetics-600:} \cite{chen2021elaborative} randomly divides the 220 new classes in Kinetics-600 into 60 validation classes and 160 testing classes, repeating this process three times. We conduct tests on the three splits provided in \cite{chen2021elaborative} and report the top-1 accuracy along with the standard deviation.

\subsubsection{Implementation Details}
The initial learning rate is set to $3.33\times10^{-6}$ and then decayed to $3.33\times10^{-8}$ using the cosine decay scheduler. We use 2 epochs to warm up the training, during which the learning rate is set to $3.33\times10^{-8}$. Subsequently, we fine-tune on Kinetics-400 for another 20 epochs. The training is executed on 8 GPUs and each contains a batch of 8 samples. During our process of vision-text alignment, augmentations like mixup and cutmix can significantly impact the alignment result. Therefore, we do not use them to prevent any potential negative effects. Instead, we simply use basic augmentations like color jittering, random flipping, and random cropping. 
Each video clip consists of 8 frames sampled with a stride of 16. During testing, we sample 3 clips with 1 crop (``$3\times1$ views'') per video to produce a prediction and aggregate the results linearly. Furthermore, we set the interval of regularization to (0.0, 0.6), the balance ratio $C$ of the regularization is set to 0.5, the balance ratio, $\gamma$ is set to 4.0 and we start SWA from the second epoch when the warm-up stage finishes. To generate video captions, we use the pretrained BLIP-2 XXL for frame captioning and use Llama-2 7B chat version for aggregating captions.

\begin{figure*}[ht]
    \centering
    \includegraphics[width=.98\textwidth]{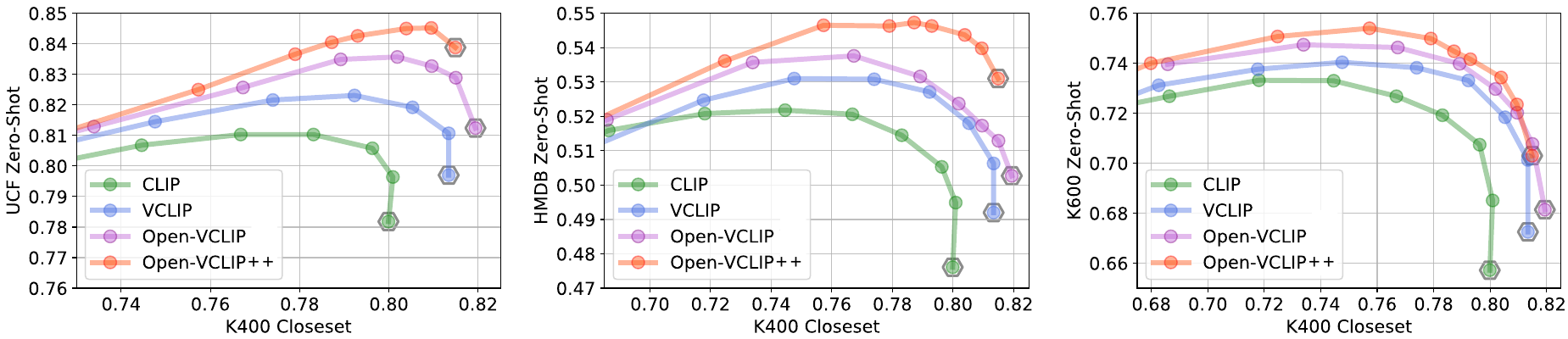}
         \caption{We evaluate the effectiveness of temporal modeling and weight interpolation with a VCLIP B/16 model. Points on each curve represent interpolation ratios in range of 0.0 to 1.0. We test on the full UCF, HMDB and the first split in~\cite{chen2021elaborative} on Kinetics-600. }
    \label{fig:discussion}
\end{figure*}

\subsection{Main Results}

\subsubsection{Comparison to state-of-the-arts action recognition}

We evaluate our method against state-of-the-art zero-shot video action recognition approaches using the UCF-101, HMDB-51, and a subset of the Kinetics-600 datasets. 
The results are presented under two distinct evaluation protocols, \textit{Evaluation Protocol 1} and \textit{Evaluation Protocol 2}, and are summarized in \cref{table:sota-compare}. The first block of \cref{table:sota-compare} presents methods that do not rely on large-scale vision-language pretrained models, while the last two blocks of \cref{table:sota-compare} demonstrate the performance of CLIP-based methods that transfer the image CLIP model to the video domain through fine-tuning on the Kinetics-400 dataset.

The table reveals that CLIP-based methods, including ActionCLIP, TEXT3VIS, and our approach, consistently outperform other methods. This indicates the significance of leveraging knowledge from large-scale vision-language pretrained models to enhance zero-shot video action recognition. Furthermore, when using the ViT-B/16 encoder, our method outperforms ActionCLIP by a substantial margin of 14.6\% (84.1\% versus 69.5\%) and TEXT4VIS by 4.2\% (54.7\% versus 50.5\%). When utilizing the ViT-L/14 encoder, our method surpasses TEXT4VIS by 8.5\% (88.1\% versus 79.6\%) and 8.9\% (58.7\% versus 49.8\%). These notable accuracy improvements highlight the effectiveness of our approach in adapting the CLIP model to the video domain for improved zero-shot recognition.

We also compare our method with X-CLIP \cite{ni2022expanding} under \textit{Evaluation Protocol 3} on the UCF and HMDB datasets. The results presented in \cref{table:protocol3} demonstrate that our approach outperforms X-CLIP by a substantial margin of 11.9\% (83.9\% versus 72.0\%) on UCF and 11.0\% (55.6\% versus 44.6\%) on HMDB when employing the ViT-B/16 encoder, suggesting the effectiveness of our approach.

When considering the Kinetics-600 dataset, as illustrated in \cref{table:zsl-kinetics-600}, our method consistently demonstrates strong performance. This is especially notable when we compare it to X-CLIP \cite{ni2022expanding} and TEXT4VIS \cite{wu2022transferring}, both of which provide a fair basis for comparison with our approach.

\begin{table}[t!]
\caption{Comparing our method with various alternative methods using different backbone networks, including ViT-B/32, ViT-B/16 and ViT-L/14.  ``CLIP$^{*}$'' denotes the results of directly applying the image pretrained CLIP model. ``FINE-TUNE'' denotes the standard fine-tuning on Kinetics-400 action recognition dataset using the same model as \system.}
    \vspace{-0.1in}
\label{table:baseline-compare}
\resizebox{.48\textwidth}{!}{
\begin{small}
\begin{sc}
\begin{tabular}{lc|>{\columncolor{white}}c|ccc}
\toprule
Net  & Dset&  CLIP$^*$ &  Fine-tune & Ours\\
\midrule
\makecell{\multirow{3}{*}{B/32}} & UCF & 69.1 & 78.0\rise{8.9} & \,\,\textbf{79.9\rise{10.8}} \\
& HMDB & 45.4 & 47.3\rise{1.9} & \textbf{51.1\rise{5.7}} \\
& K600 & 64.8 & 62.8\drop{2.0} & \textbf{69.5\rise{4.7}} &  \\
\midrule
\makecell{\multirow{3}{*}{B/16}} & UCF & 74.2 & 79.7\rise{5.5}  & \textbf{84.1\rise{9.9}} & \\
& HMDB & 47.6 & 49.2\rise{1.6} & \textbf{54.7\rise{7.1}} \\
& K600 & 68.1 & 65.9\drop{2.2} & \textbf{73.4\rise{5.3}} \\
\midrule
\makecell{\multirow{3}{*}{L/14}} & UCF &  80.5 & 85.0\rise{4.5} & \textbf{88.1\rise{7.6}} \\
& HMDB & 55.0 & 51.9\drop{3.1} & \textbf{58.7\rise{3.7}} \\
& K600 & 76.2 & 74.9\drop{1.3} & \textbf{81.2\rise{5.0}} \\
\bottomrule
\end{tabular}
\end{sc}
\end{small}
}
\end{table}

\subsubsection{Results with different backbones}

We compare \system  with the CLIP baseline and standard fine-tuning using various backbone networks. In particular, the results in the ``CLIP$^*$'' column of \cref{table:baseline-compare} show the performance of the pretrained CLIP model, which predicts each image frame of the video seperately. The ``FINE-TUNE'' column shows the results of a standard fine-tuning process using the same model as \system. We see from \cref{table:baseline-compare} that larger backbone networks (ViT-L/14 $>$ ViT-B/16 $>$ ViT-B/32) consistently improve zero-shot performance. For example, the results in the ``CLIP$^*$'' column show that the ViT-L/14 model outperforms the ViT-B/16 by 6.3\% (80.5\% \vs 74.2\%), 7.4\% (55.0\% \vs 47.6\%), and 8.1\% (76.2\% \vs 68.1\%) on the UCF, HMDB, and Kinetics-600 datasets, respectively, while similar trends can be found in the last two columns of \cref{table:baseline-compare}, demonstrating that larger-scale vision-language pretrained models contain stronger zero-shot knowledge and are more robust when transferred from image to video tasks. However, as we can see from the results of ``FINE-TUNE'' and ``CLIP$^*$'' in \cref{table:baseline-compare}, the fine-tuning process does not always result in improved performance. For example, the fine-tuning process leads to worse zero-shot results on the HMDB dataset when using the ViT-L/14 backbone, and all zero-shot testing results on the Kinetics-600 show a performance drop after models are fine-tuned on Kinetics-400 under a standard paradigm.  
The results indicate that the knowledge stored in CLIP cannot always be transferred with conventional fine-tuning.

Compared to the baselines, our approach demonstrates significant and consistent improvements in zero-shot video action recognition across various datasets and models, as shown in the last column of Table~\ref{table:baseline-compare}. Taking experimental results on Kinetics-600 as an example, our approach not only avoids the performance drop seen in the ``FINE-TUNE'' method, but also significantly improves zero-shot recognition. This highlights the effectiveness of our method in effectively using annotated video datasets with limited labels for transfer learning and effectively preventing knowledge from forgetting, offering stable and decent results.

\begin{figure*}[ht!]
    \centering
    \includegraphics[width=0.96\textwidth]{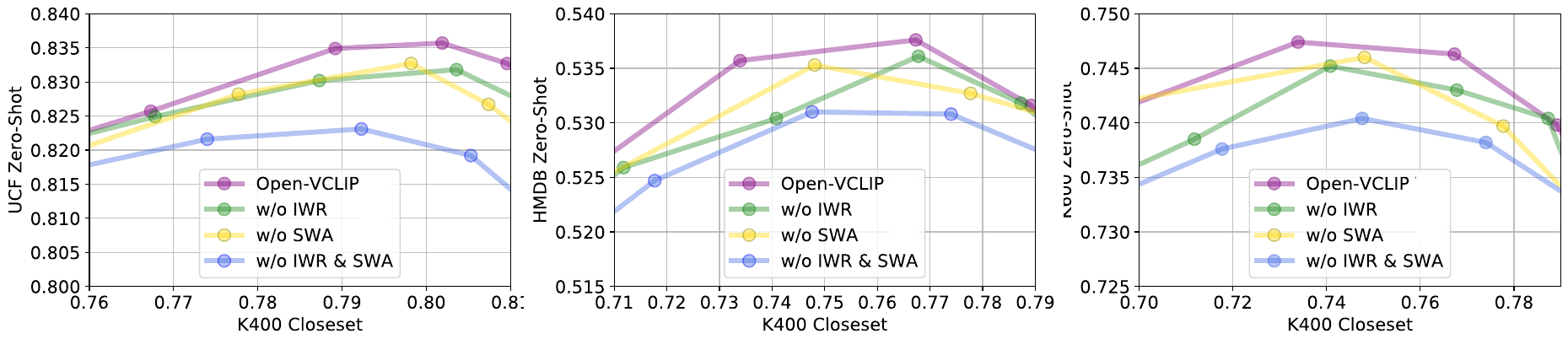}
     \caption{We evaluate the effectiveness of IWR and SWA with a VCLIP B/16 on the full UCF, HMDB and the first split in~\cite{chen2021elaborative} on Kinetics-600.} 
    \label{fig:ablation}
\end{figure*}

\begin{figure*}[ht!]
    \centering
    \includegraphics[width=0.96\textwidth]{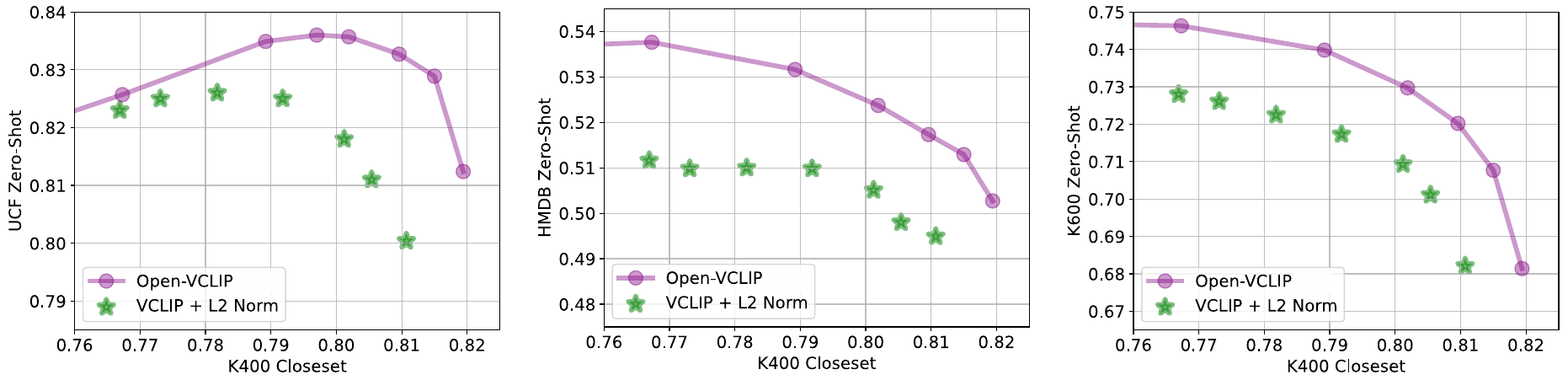}
     \caption{We compare \history with an $\ell_2$ regularization imposed on weigths on the full UCF, HMDB and the first split in~\cite{chen2021elaborative} on Kinetics-600.} 
    \label{fig:l2_regularization}
\end{figure*}

\subsection{Discussion}

\subsubsection{The effectiveness of different components}

We first investigate the contributions of different components,
 \ie, temporal modeling, weight interpolation, and text generation, for transferring CLIP to the video domain. To this end, we compare the performance of the CLIP, VCLIP, \history and \system, represented by green, blue, purple and red curves in \cref{fig:discussion}, respectively. In particular, VCLIP expands the original spatial attention of CLIP to spatial-temporal attention without additional parameters as in \history and \system. \history applies the interpolated weights regularization and stochastic weight averaging based on VCLIP. Based on \history, \system further expands the original narrow text space of Kinetics-400 with different kinds of captions extracted by the combination of BLIP-2 and Llama-2. The curves illustrate weight interpolations with different mixing coefficients between CLIP and the fine-tuned model. Concretely, the $y$-axis displays the accuracy on the zero-shot video dataset, while the $x$-axis displays accuracy on the fine-tunning dataset, Kinetics-400.

Comparing the tails of the green and blue curves, we see that VCLIP achieves not only better close-set performance, but also better zero-shot performance on all of the three datasets. At the same time, when applying weight interpolation with different ratios, VCLIP achieves better trade-off than CLIP, as evidenced by the fact that the blue curve is always on top of the green curve in \cref{fig:discussion}. 
This strongly suggests the significant advantages that come with incorporating temporal modeling in the context of zero-shot video action recognition.

Furthermore, the clear margins between the purple curves and the blue curves in \cref{fig:discussion} show that our proposed weight interpolation helps the model achieve better trade-off compared to pure fine-tuning. With the same-level of close-set results, \history always produces better zero-shot accuracy. Finally, the improvements observed between the red curves and the other curves in \cref{fig:discussion} show that our final proposed solution achieves the best trade-off, thus demonstrating the importance of using additional textual data for better alignment.

\begin{table*}[ht!]
\caption{Comparison of zero-shot video-to-text/text-to-video retrieval performance for various algorithms. ``T2VRN'' denotes the recall@N of text-to-video retrieval. ``V2TRN'' denotes the recall@N of video-to-text retrieval. Many of the current methods pretrain models on a vast array of datasets. Here, INET21k denotes to ImageNet-21k~\cite{deng2009imagenet} and INET1k denotes to ImageNet-1k~\cite{deng2009imagenet}. CC3M~\cite{sharma2018conceptual}, CC12M~\cite{changpinyo2021conceptual} and SBU captions~\cite{ordonez2011im2text} are three web datasets. WV-2M and WV-10M are the WebVid datasets proposed in \cite{bain2021frozen}. HT100M is HowTo100M~\cite{miech2019howto100m}. COCO~\cite{lin2014microsoft} and VG(Visual Genome)~\cite{krishna2017visual} are two human-annotated datasets. LAION-100M~\cite{schuhmann2021laion}, WIT-400M~\cite{radford2021learning} are web-scale image-text pairs dataset. SS-V2~\cite{goyal2017something}, K400~\cite{kay2017kinetics} and K710~\cite{li2022uniformerv2} are action recognition datasets. SC-V denotes to the self-collected video dataset in ~\cite{wang2022internvideo}.}
\label{table:vl-retrieval}
\begin{center}
\begin{sc}
\begin{tabular}{@{\extracolsep{\fill}\quad\quad}lcccccccccc}
\toprule
Method & dataset & backbone & t2vR1 & t2vR5 & t2vR10 & v2tR1 & v2tR5 & v2tR10  \\
\midrule
\multicolumn{3}{l}{\small{\textit{non CLIP-based methods}}} \\
Frozen\cite{bain2021frozen} & INet21k,CC3M+WV-2M,COCO & ViT-B/16 & 24.7 & 46.9 & 57.2 & - & - & - \\
\cmidrule(lr){2-2}
OmniVL\cite{wang2022omnivl} & \makecell{\text{INet1k,CC3M,CC12M,WV-2M,}\\\text{COCO,VG,SBU,K400}} & ViT-B/16 & 34.6 & 58.4 & 66.6  
& - & - & -\\
\cmidrule(lr){2-2}
SINGULARITY\cite{lei2022revealing} & \makecell{\text{INet21k,CC3M,CC12M,WV-2M,}\\\text{COCO,VG,SBU}} & ViT-B/16& 34.0 & 56.7 & 66.7 & - & - & - \\ 
\midrule 
\multicolumn{3}{l}{\small{\textit{CLIP-based methods (with WIT-400M dataset)}}} \\
CLIP\cite{portillo2021straightforward} & - &  ViT-B/16 & 31.2 & 53.7 & 64.2 & 27.2 & 51.7 & 62.6 \\
\cmidrule(lr){2-2}
CLIP4CLIP\cite{luo2022clip4clip} & HT100M-380k & ViT-B/16 & 32.0 & 57.0 & 66.9 & - & - & - \\
\cmidrule(lr){2-2}
InternVideo\cite{wang2022internvideo} & \makecell{\text{LAION-100M,WV-2M,WV-10M,}\\\text{HT100M,AVA,SS-V2,}\\\text{SC-V,K400,K710}} & ViT-L/14 & 40.7 & - & - & 39.6  & - & - \\
\midrule 
\makecell{\multirow{2}{*}{\system}} & \makecell{\multirow{2}{*}{K400}} & ViT-B/16 & 34.6 & 59.8 & 69.5 & 35.4 & 60.3 & 72.0\\
 & & ViT-L/14 & 39.0 & 63.3 & 72.3 & 37.8 & 62.7 & 73.0\\
\bottomrule
\end{tabular}
\end{sc}
\end{center}
\end{table*}

\subsubsection{IWR \vs SWA}

Our training algorithm is composed of two weight interpolation modules: IWR which is used during training to regularize the fine-tuning process and SWA to improve generalization. We investigate their contributions to the final results in \cref{fig:ablation}. Overall, removing either IWR (green curve) or SWA (yellow curve) from the model leads to significant drops, \ie the red curve outperforms all other curves, suggesting IWR and SWA are complimentary to each other.
Furthermore, we see using only IWR or SWA is able to produce a good zero-shot performance improvement, compared to the results with no interpolation at all. 

\subsubsection{Weight regularization during fine-tuning}

IWR is conceptually similar to EWC~\cite{ewc}, which penalizes the changes of important parameters for solving previous tasks. However, EWC needs to assess the importance of parameters through historical data which is not feasible to our setting. Instead, we simply 
constrain the optimization by penalizing the weight changes using an $\ell_2$ norm.
 
Concretely, the training process introduces a regularization loss term derived from the $\ell_2$ distance between updated and original weights. This regularization term is assigned varying weights to fine-tune VCLIP on the original Kinetics-400 dataset. As shown in \cref{fig:l2_regularization}, the diverse green data points correspond to outcomes with varying $\ell_2$ penalization ratios, while the purple curves depict fine-tuning VCLIP with our proposed weight interpolated optimization. We can see our method achieves the best trade-off. \history achieves higher close-set and zero-shot accuracy across all the datasets compared with simply applying an $\ell_2$ normalization, demonstrating the effectiveness of \history.  It is worth noting that our proposed approach only needs to train the model once and then obtain models with different trade-offs by modifying the interpolated weight coefficients, whereas the method employing $\ell_2$ norm penalization requires to retrain the full model each time when the regularization coefficient is altered, leading to a significantly heightened training cost.

\subsubsection{Comparison with parameter-efficient fine-tuning}

Fine-tuning only a part of the weights of a network is an effective approach for adapting the CLIP model to video in a parameter-efficient manner. A noteworthy method that employs this strategy is the ST-Adapter~\cite{pan2022st}, which accomplishes effective video classification by training added adapter modules, while keeping the original parameters of the CLIP model frozen. We compare ST-Adapter with our approach.
More specifically, ST-Adapter forgoes the text encoder of CLIP, while introducing an additional linear classifier instead. This change prevents the model from being used in zero-shot testing. To circumvent this problem, we incorporate the text encoder of CLIP into our implementation of the ST-Adapter. The results presented in Table~\ref{table:compare_stadapter} illustrates that the ST-Adapter, while preserving the original CLIP weights and only fine-tuning the added adapters, fails to match the performance of our proposed method in zero-shot action recognition. In particular, we observe a marked degradation in the zero-shot performance of the ST-Adapter on the K600 dataset, suggesting that parameter-efficient fine-tuning does not effectively address the issue of catastrophic forgetting.

\begin{table}[h]
\caption{Comparison with parameter efficient fine-tuning method. }
\label{table:compare_stadapter}
\begin{center}
\begin{small}
\begin{sc}
\begin{tabular}{l|cccccccccc}
\toprule 
 Method & UCF & HMDB & K-600 \\
\midrule
CLIP & 74.2 & 47.6 & 68.1$\pm$1.0 \\
Fine-tuned VCLIP & 79.7 & 49.2 & 65.9$\pm$1.0 \\
ST-Adapter & 77.3 & 49.8 & 60.2$\pm$1.8 \\
\midrule
\system  & 84.1 & 54.7 & 73.4$\pm$0.8 \\
\bottomrule  
\end{tabular}
\end{sc}
\end{small}
\end{center}
\end{table}

\begin{figure*}[t]
    \centering
    \includegraphics[width=0.95\textwidth]{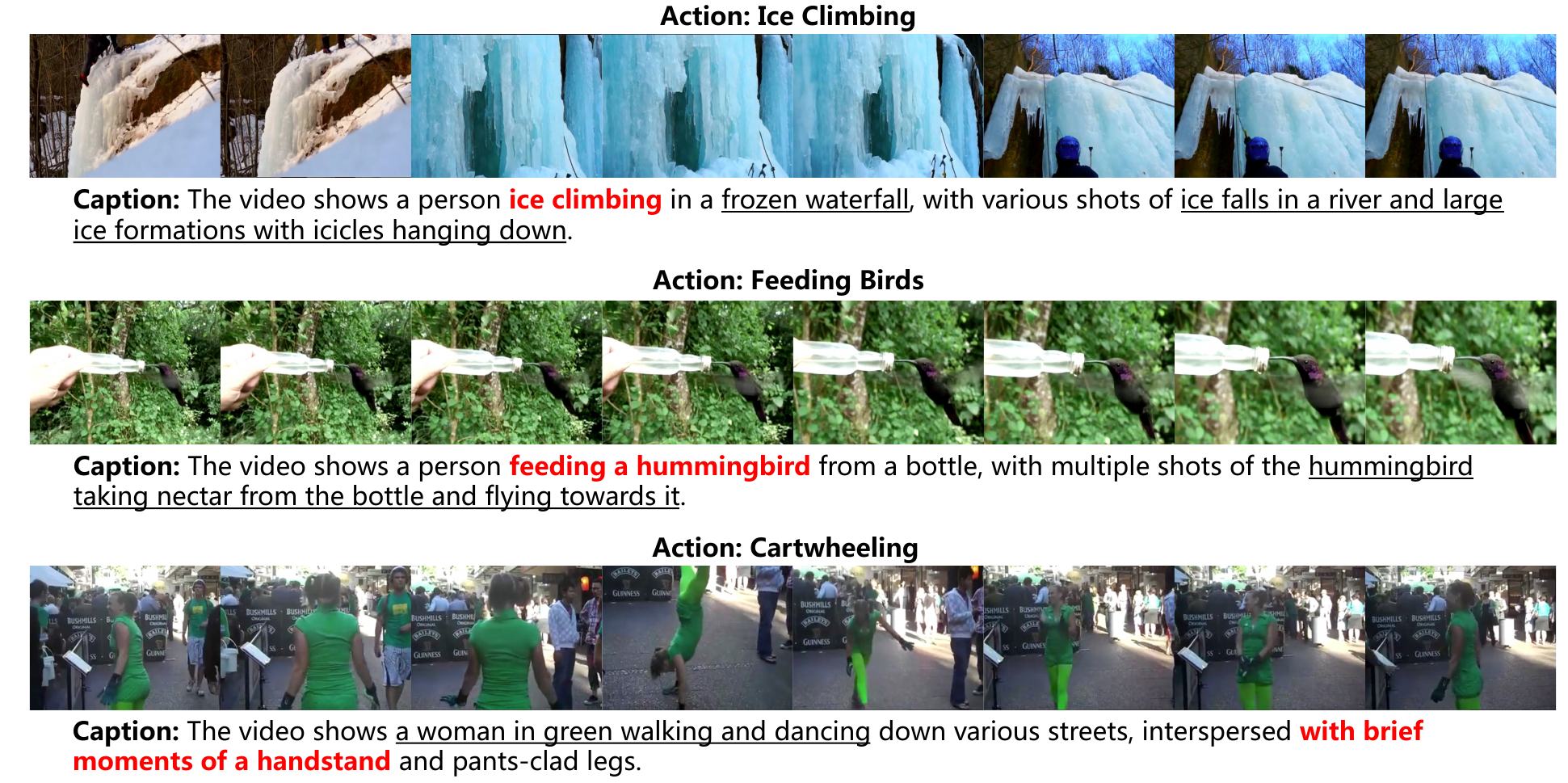}
    \caption{{ We show captions generated by our BLIP-2 + Llama-2 pipeline. The  content in the caption is highlighted in red if it correctly corresponds to the video action; otherwise, it is highlighted in blue. The additional detail descriptions are underlined.}.  }
    \label{fig:showcase}
\end{figure*}

\begin{table*}[h!]
\centering
  \caption{{We adjust the loss ratio of vision-text alignment to explore the importance of text space expansion. The top 1 and top 2 score of each column are colored in red and blue, respectively.}}
    \vspace{-0.1in}
  \label{tab:capadjust}
  \begin{sc}
  \begin{tabular}{c||c|ccc|cccccc}
    \toprule
    \multirow{2}{*}{Exp} & \multicolumn{1}{c|}{Component} & \multicolumn{3}{c|}{Zero-shot Classification} & \multicolumn{6}{c}{Zero-shot Retrieval}  \\
    \cmidrule(lr){2-2}\cmidrule(lr){3-5}\cmidrule(lr){6-8}\cmidrule(lr){9-11}
    &   Cap Ratio & UCF & HMDB & K-600 & t2vR1 & t2vR5 & t2vR10 & v2tR1 & v2tR5 & v2tR10 \\
    \midrule 
    \multirow{4}{*}{Full} &   0.0 & 83.5&53.2&73.0& 31.3 & 54.3 & 65.6 & 33.6 & 59.1 & 70.0 \\
    \cmidrule(lr){2-11}
    & 1.0 & 83.9 & 53.9 & 73.2 & 33.0 & 57.3 & 68.9 & 34.7 & 58.6 & 70.4 \\
    & 2.0 & 83.9 & 54.3 & \color{red}{\textbf{73.5}} & 33.2 & 57.9 & 69.1 & \color{red}{\textbf{35.5}} & 59.3 & 70.9 \\
    & 3.0 & \color{blue}{84.0} & \color{blue}{54.6} & \color{red}{\textbf{73.5}} & \color{blue}{33.9} & \color{blue}{58.9} & \color{blue}{69.2} & 35.0 & \color{red}{\textbf{60.6}} & \color{blue}{71.3} \\
    & 4.0 & \color{red}{\textbf{84.1}} & \color{red}{\textbf{54.7}} & \color{blue}{73.4} & \color{red}{\textbf{34.6}} & \color{red}{\textbf{59.8}} & \color{red}{\textbf{69.5}} & \color{blue}{35.4} & \color{blue}{60.3} & \color{red}{\textbf{72.0}} \\
    \bottomrule
  \end{tabular}
  \end{sc}
\end{table*}

\subsubsection{Text-to-Video/Video-to-Text Retrieval Performance}
Assessing the model via text-to-video/video-to-text retrieval tasks offers insights into its generalizability within the video domain. We follow the paradigm of training models on Kinetics-400 dataset and testing them on MSR-VTT dataset~\cite{xu2016msr}, which is a large video description dataset. 
{\color{black} We compare our approach with a variety of video-language alignment methods using the standard retrieval metric: recall at rank K (Recall@K), which is the same as in~\cite{bain2021frozen}. T2V and V2T in~\cref{table:vl-retrieval} denotes to text-to-video and video-to-text retrieval results respectively, while R1, R5 and R10 denotes to Recall@1, Recall@5 and Recall@10. We also report the training dataset, network architecture used for better comparisons.}

First, We report the result of our own implementation using CLIP for retrieval as shown in the first row of the CLIP-based methods block in~\cref{table:vl-retrieval}, which is similar to the results in~\cite{portillo2021straightforward}. This guarantees fair comparisons. Compared to using the original CLIP, \system significantly improves both the text-to-video and video-to-text retrieval recalls, \eg, the text-to-video Recall@1, Recall@5, Recall@10 surpass the CLIP baseline by 3.4\% (34.6\% \vs 31.2\%), 6.1\% (59.8\% \vs 53.7\%) and 5.3\% (69.5\% \vs 64.2\%), while the video-to-text Recall@1, Recall@5, Recall@10 surpass the CLIP baseline by 8.2\% (35.4\% \vs 27.2\%), 8.6\% (60.3\% \vs 51.7\%) and 9.4\% (72.0\% \vs 62.6), demonstrating the effectiveness of our method in preserving the alignment capability when transferring to video domain.

Besides, \system is much more training efficient compared to  current polular methods. As shown in ~\cref{table:vl-retrieval}, we find that current researchers typically improve the  zero-shot retrieval performance by pretraining on a vast array of datasets. For example, OMNIVL~\cite{wang2022omnivl} trains models on 7 more datasets in addition to ImageNet, CLIP4CLIP~\cite{luo2022clip4clip} fine-tunes the CLIP-based model on HowTo100M-380k dataset, and InternVideo~\cite{wang2022internvideo} fine-tunes the CLIP-based model on 9 more large datasets in addition to WIT-400M~\cite{radford2021learning}. Instead, \system trains CLIP models only on Kinetics-400 and achieves comparable or better results. For instance, compared with CLIP4CLIP which fine-tunes CLIP on more video-text pairs, \system achieves better text-to-video performance; compared with OMNIVL which is trained on not only Kinetics-400, but also 7 more datasets, \system offers better text-to-video recalls, \eg Recall@5: 59.8\% \vs 58.4\% and Recall@10: 69.5\% \vs 66.6\%; even compared with InternVideo that uses a substantial amount of training data (not only using the pretraining weights of CLIP, but also introducing a very wide variety of video or image data during training), \system achieves the comparable results with  much fewer data.

\subsubsection{The importance of alignment with pseudo captions}

\system additionally align visual and textual modalities with additional captions compared to the preliminary version \history. To ablate the effectiveness of doing so, we report the results of \system using different ratios for the alignment loss.

We observe in~\cref{tab:capadjust} that the zero-shot performance on action classification and video-text retrieval tasks are improved in almost all cases by using additional captions. We also see that larger ratios that inject stronger supervision from captions bring more performance improvements, demonstrating the effectiveness of such captions. 
Besides, we find aligning models on the additional captions obviously benefits more for the video-text retrieval tasks. For instance, as shown in~\cref{tab:capadjust}, text-to-video Recall@1, Recall@5 and Recall@10 get improved by 3.3\% (34.6\% \vs 31.3\%), 5.5\% (59.8\% \vs 54.3\%) and 3.9\% (69.5\% \vs 65.6\%) respectively, and the performance gain is more obvious compared to the performance gain on the action recognition.

\subsubsection{Examples of generated captions  }

\cref{fig:showcase} shows examples of captions 
generated by the combination of the BLIP-2 model and the Llama-2. It can be seen that more video details are included in the captions compared to the original action labels of the videos, \eg, in the first video, the new caption description not only contains the action information: ``ice climbing'', but also describes in detail about the environments: ``frozen waterfall'', ``ice falls in a river and large ice formations with icicles hanging down''. Also in the second video, we gain the more detailed description ``feeding a hummingbird'' compared to the original action label ``feeding birds''. However, we also find the generated captions are not always perfect. For example, as we can see in the third video, 
although the caption is more informative with the description ``a woman in green walking and dancing'', there may be a small discrepancy in the action description, where the action of ``cartwheeling'' is described as ``with brief moments of a handstand''.

\subsubsection{Results of \system on ImageNet}

\begin{figure}[t]
    \centering
    \includegraphics[width=0.4\textwidth]{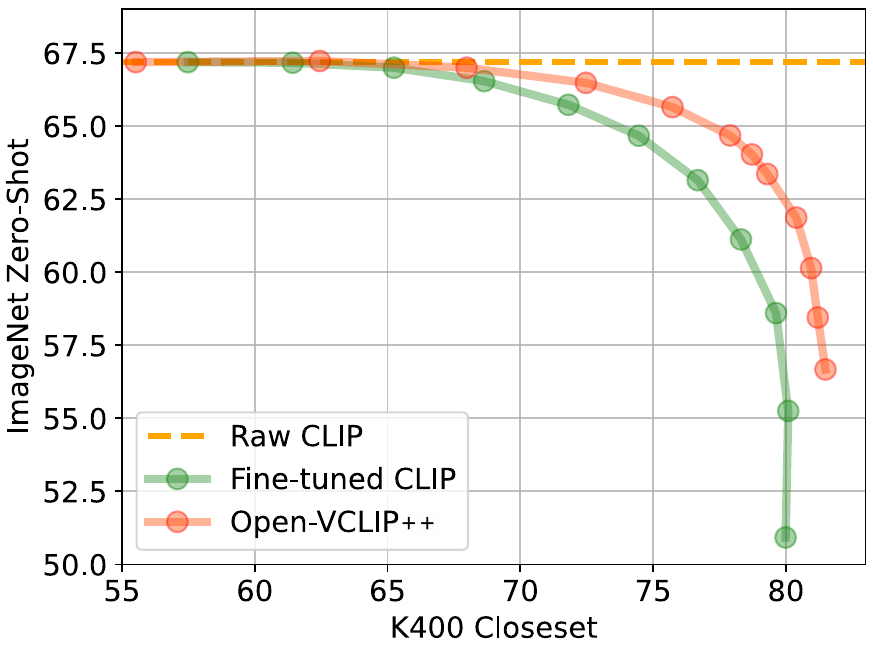}
    \caption{{We report the performance trade-off between the close-set performance on K400 and the zero-shot performance on ImageNet. Notably, the fine-tuned \system can be tested on the image task by simply turning off the extended attention view operation.} }
    \label{fig:showforget}
\end{figure}

We conduct additional experiments on image classification~\cref{fig:showforget} to demonstrate our method is better able to maintain the original knowledge of CLIP for image tasks as well.
Notably, \system excels in achieving a decent trade-off on the ImageNet dataset, thereby showcasing its proficiency in mitigating knowledge forgetting. This demonstrates a remarkable ability of \system to adapt to downstream tasks while better retaining previous knowledge.

\section{Conclusion}

We presented \system, an effective approach that enables CLIP to be transformed to an open-vocabulary video model. \system contains lightweight temporal modeling modules that equip CLIP with the ability to capture spatial and temporal relationships in videos. More importantly, \system is optimized with a carefully designed regularization strategy that strives for generalization to preserve the zero-shot abilities of CLIP. Furthermore, we propose a video captioning strategy to expand the text space, so as to reduce the risk of overfitting.
Extensive experiments are conducted and the results demonstrate that \system outperforms state-of-the-art methods with clear margins on zero-shot video action recognition and achieves the best trade-off between close-set and zero-shot video action recognition. One potential limitation is that adversaries could craft membership inference attacks to steal information from the model.

\bibliography{main}
\bibliographystyle{IEEEtran}

 \begin{IEEEbiography}[{\includegraphics[width=1.2in,height=1.25in,clip,keepaspectratio]{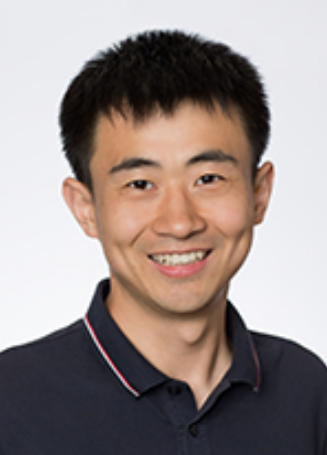}}]{Zuxuan Wu} 
 received his Ph.D. in Computer Science from the University of Maryland in 2020. He is currently an Associate Professor in the School of Computer Science at Fudan
 University. His research interests are in computer vision and deep learning. His work has been recognized by an AI 2000 Most Influential Scholars Award in 2022, a Microsoft Research PhD Fellowship in 2019 and a Snap PhD Fellowship in 2017. \end{IEEEbiography}

 \begin{IEEEbiography}[{\includegraphics[width=1.2in,height=1.25in,clip,keepaspectratio]{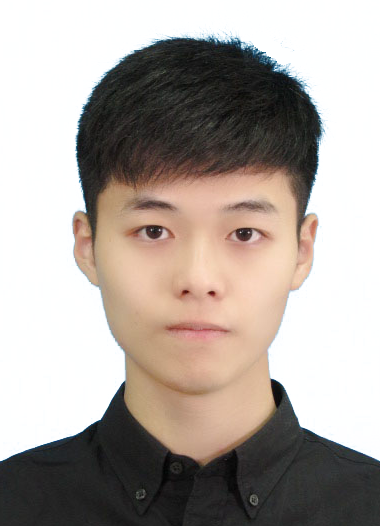}}]{Zejia Weng} received the B.S. degree in 2020 from
Fudan University, Shanghai, China, where he is currently working toward the Ph.D. degree with the
School of Computer Science. His research interests
include computer vision, deep learning, and especially large-scale video understanding.
 
 \end{IEEEbiography}

  \begin{IEEEbiography}[{\includegraphics[width=1.2in,height=1.25in,clip,keepaspectratio]{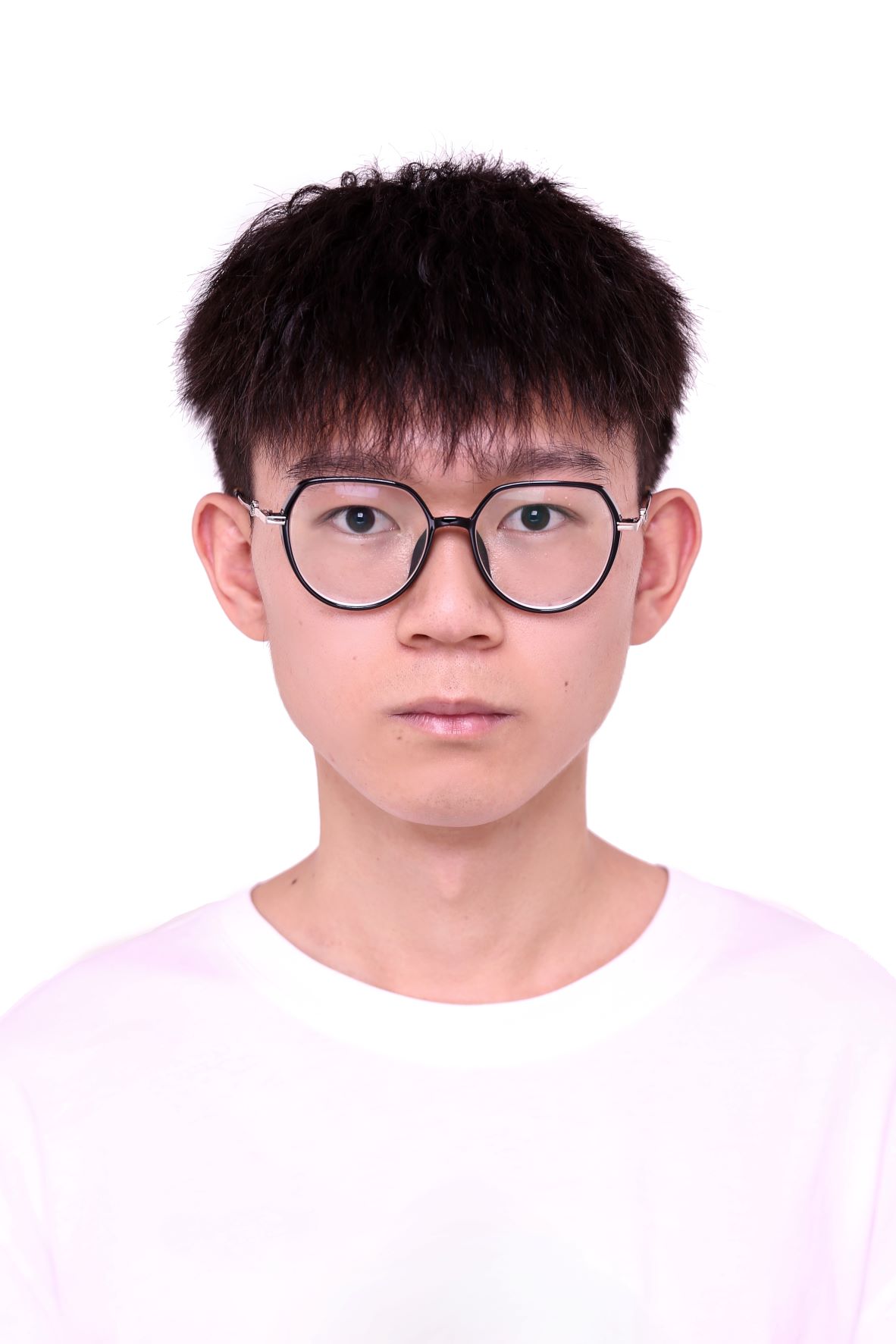}}]{Wujian Peng} received his B.S. degree in Software Engineering from Xi'an Jiaotong University in 2022. He is currently a second-year graduate student at the School of Computer Science, Fudan University. His research interests lie in computer vision, especially on semi-supervised learning and vision-language modeling.
 \end{IEEEbiography}

\begin{IEEEbiography}[{\includegraphics[width=1in,height=1.25in,clip,keepaspectratio]{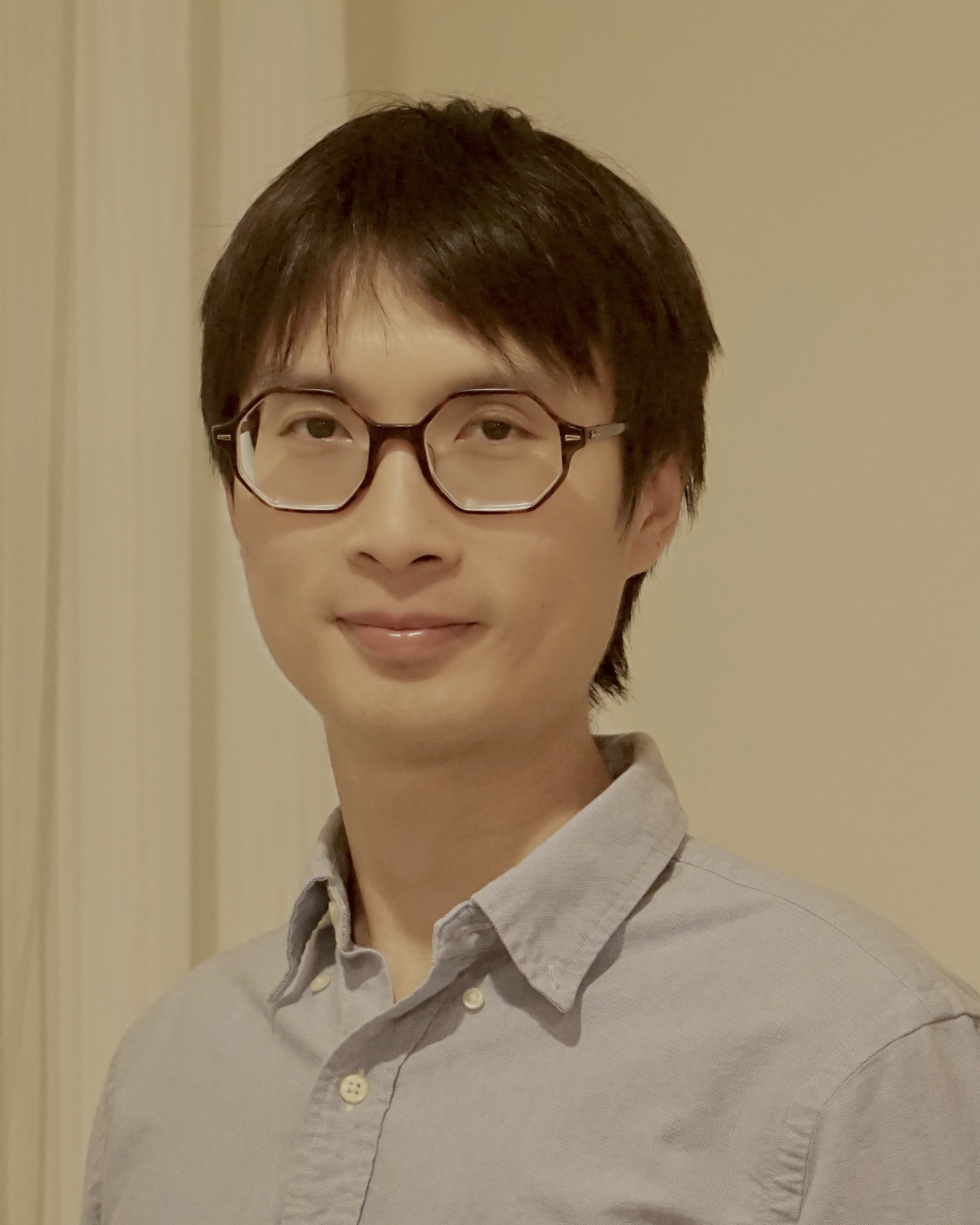}}]
  {Xitong Yang} received the B.E. degree from Beijing Institute of Technology in 2014, the M.S degree from University of Rochester in 2016, and the Ph.D. degree from University of Maryland, College Park in 2021. He is currently a research scientist at Meta AI. His research interest covers computer vision, deep learning and machine learning, with specific focus on large-scale, long-form video understanding.
\end{IEEEbiography}

\begin{IEEEbiography}[{\includegraphics[width=1in,height=1.25in,clip,keepaspectratio]{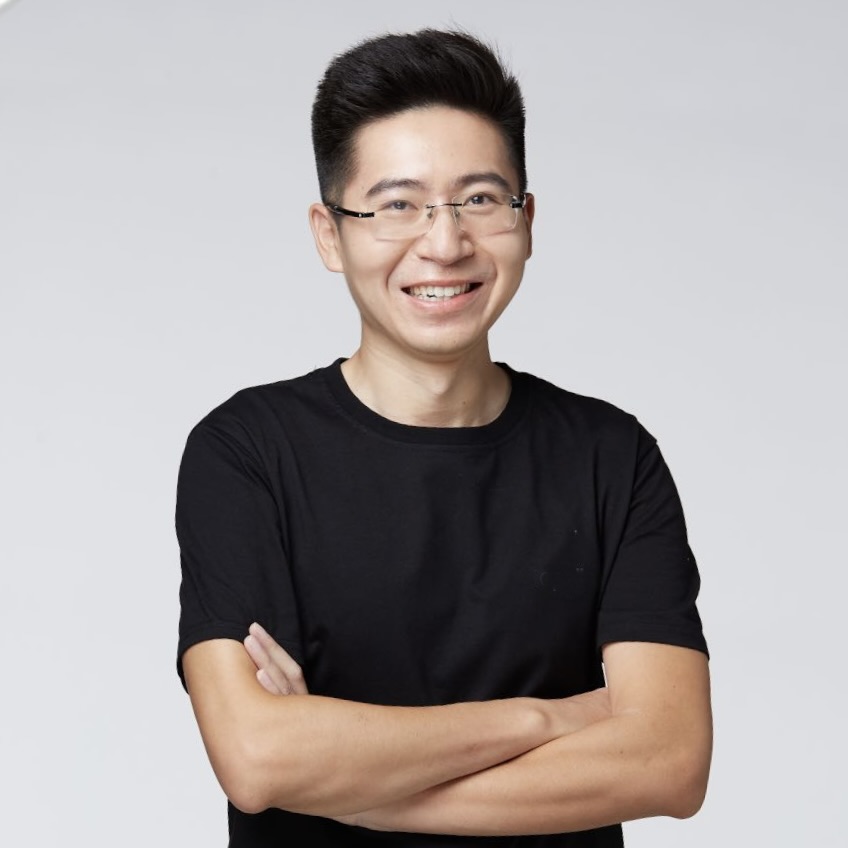}}]
  {Ang Li} is CEO and Co-Founder at Simular. He received my Ph.D. in computer science from the University of Maryland College Park in 2017 and B.S. in Computer Science from Nanjing University in 2011. He was Principal Scientist and Head of Software Engineering at Baidu Apollo USA from 2021-2023, Staff Research Scientist at Google DeepMind from 2017-2021. He conducted research in Artificial Intelligence in Facebook AI Research, CMU Robotics, Apple, Google, and Comcast Labs DC.  His prior research includes various topics such as continual learning, large scale deep learning, out-of-distribution detection, vision and language, and ML infrastructure. He has served as a senior program committee member for AAAI, IJCAI, and AAMAS, as well as session chair for IJCAI 2021 and area chair for the NeurIPS workshop on Meta-Learning.
\end{IEEEbiography}

\begin{IEEEbiography}[{\includegraphics[width=1in,height=1.25in,clip,keepaspectratio]{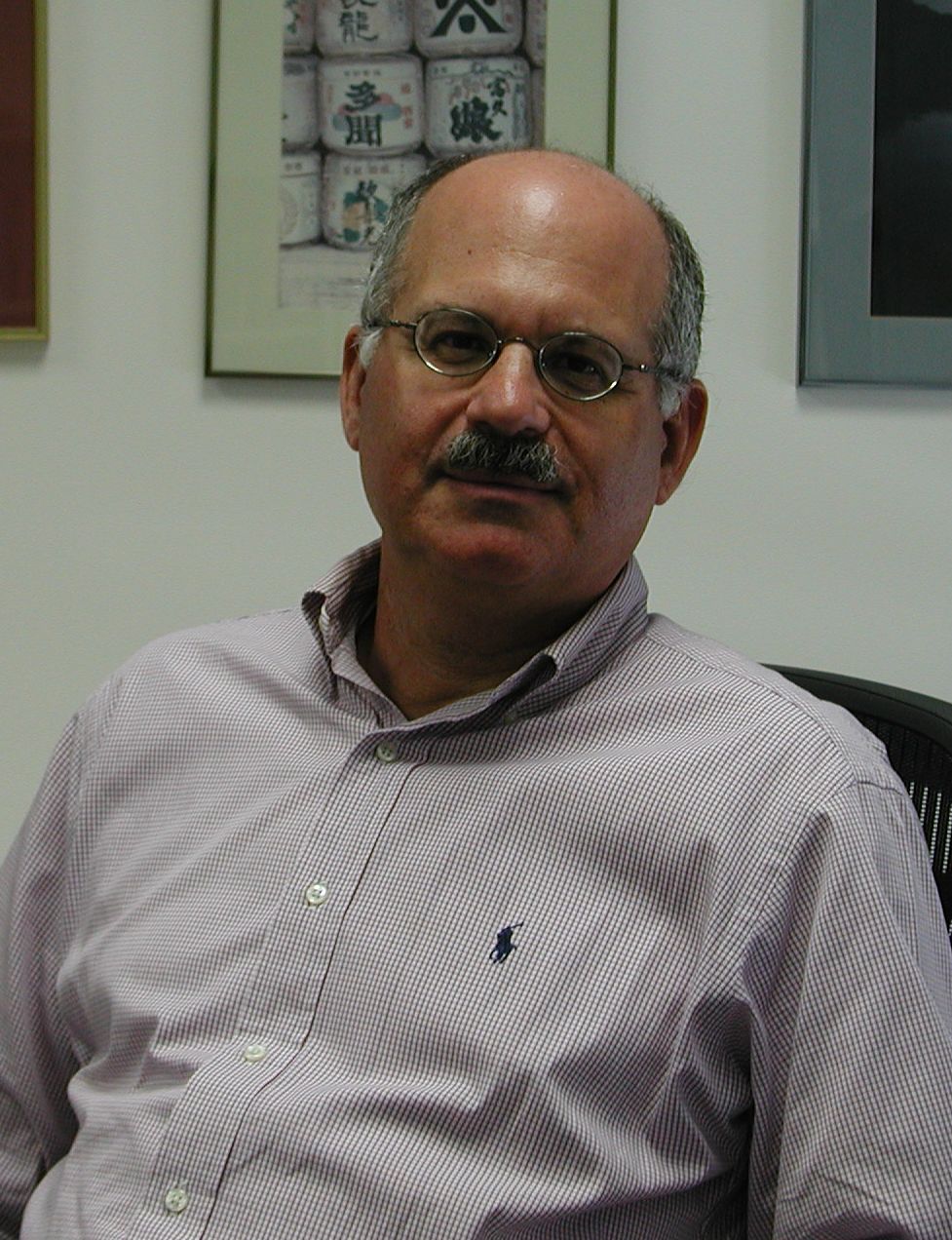}}]
  {Larry S. Davis} received the BA degree from Colgate University, in 1970 and the MS and PhD degrees in computer science from the University of Maryland, in 1974 and 1976, respectively. From 1977 to 1981, he was an assistant professor in the Department of Computer Science, University of Texas, Austin. He returned to the University of Maryland as an associate professor, in 1981. From 1985 to 1994, he was the director of the University of Maryland Institute for Advanced Computer Studies. From 1999 to 2012, he was the chair of the Computer Science Department in the institute. He is currently a professor in the institute and in the Computer Science Department. He is a fellow of the ACM, the IEEE, and the IAPR.
\end{IEEEbiography}

 \begin{IEEEbiography}[{\includegraphics[width=1.2in,height=1.25in,clip,keepaspectratio]{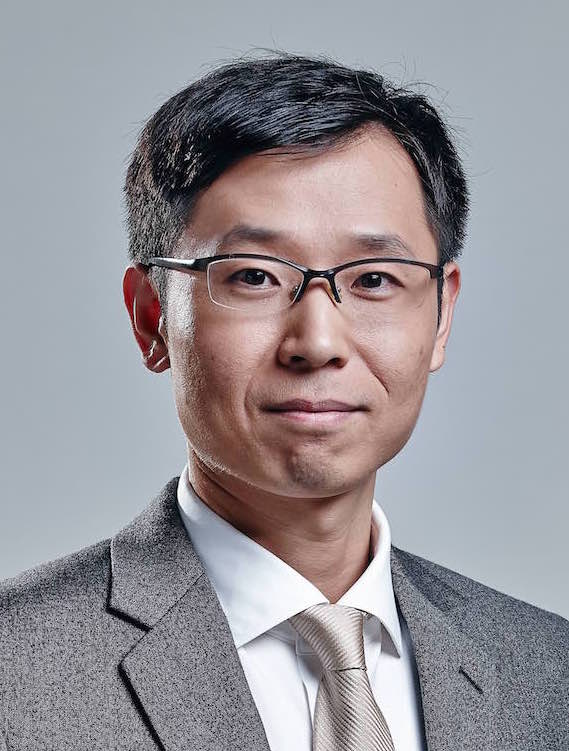}}]{Yu-Gang Jiang} received the Ph.D. degree in Computer Science from City University of Hong Kong in 2009 and worked as a Postdoctoral Research Scientist at Columbia University, New York, during 2009-2011. He is currently a Professor of Computer Science at Fudan University, Shanghai, China. His research lies in the areas of multimedia, computer vision, and robust and trustworthy AI. His work has led to many awards, including the inaugural ACM China Rising Star Award, the 2015 ACM SIGMM Rising Star Award, the Research Award for Excellent Young Scholars from NSF China, and the Chang Jiang Distinguished Professorship appointed by Ministry of Education of China. He is a fellow of IAPR.
 \end{IEEEbiography}
\vfill

\end{document}